\documentclass[journal]{IEEEtran}
\usepackage[T1]{fontenc}
\usepackage{multicol}
\usepackage{times,amsmath,amssymb,booktabs,array,graphicx,cite,url,xcolor,stfloats,placeins}
\usepackage[hidelinks]{hyperref}
\makeatletter
\newcommand{\inlinetablecaption}[1]{\def\@captype{table}\caption{#1}}
\newcommand{\methodfigurecaption}[1]{\def\@captype{figure}\caption{#1}}
\makeatother

\title{PhraseSAM: Phrase Selection and Class Correction for Open-Vocabulary Remote Sensing Segmentation}
\author{Zining~Zhu, Shukai~Yang, Haoxuan~Li, Yi~Sun, Junbo~Ba, and Xiuwei~Zhang%
\thanks{\textit{(Corresponding author: Xiuwei Zhang.)}}%
\thanks{The authors are with the School of Computer Science and Engineering, Northwestern Polytechnical University, Xi'an 710072, China (\mbox{e-mail:} zhuzining@mail.nwpu.edu.cn, yangshukai123@mail.nwpu.edu.cn, li\_haoxuan@mail.nwpu.edu.cn, 2020264819@mail.nwpu.edu.cn, bajunbo@mail.nwpu.edu.cn, and xwzhang@nwpu.edu.cn).}}
\begin{document}
\raggedbottom
\maketitle
\begin{abstract}
Open-vocabulary semantic segmentation uses text phrases to flexibly specify categories of interest across diverse remote-sensing tasks. However, the same class can have very different appearances across remote-sensing scenes. A single phrase may therefore cover only part of the class. Although expanding the phrase set can improve coverage, some phrases repeat existing responses, while others may activate competing classes and alter pixel labels. Semantically similar classes can still compete for the same pixels after phrase selection. We propose PhraseSAM, a training-free framework built on frozen SAM 3 that selects complementary phrases and corrects class conflicts. A large language model directly generates foreground candidate phrases and a background phrase set from the full class list. Foreground phrases are selected for class support, separation from competing classes, and complementary coverage; the background phrase set remains fixed. PhraseSAM identifies competing classes from their mutual responses and uses calibration pixels to set thresholds for changing pixel labels between those classes. The selected phrase set and correction rules are fixed before inference, with at most one correction per pixel. Across eight remote-sensing datasets, PhraseSAM achieves 60.5\% mean mIoU, compared with 54.2\% for GLoP-SAM3.
\end{abstract}

{\small\noindent The code is available at \url{https://github.com/crosage/phrasesam}.\par}


\section{Introduction}
Open-vocabulary semantic segmentation uses semantic associations between vision and language to identify categories specified by text phrases and assign them to image pixels. By changing the text queries, it can recognize semantic concepts beyond a predefined category set. The same remote-sensing model can therefore support different mapping tasks through text queries. For example, these queries can specify broad land-cover classes for regional surveys or distinguish roofs from facades in urban mapping. Different mapping tasks require different land-cover and object categories, making it difficult to obtain pixel-level annotations for every region and target category. Specifying categories of interest through text phrases broadens the applicability of segmentation models across mapping tasks.

Existing approaches include trained and training-free methods. SAN~\cite{san} and CAT-Seg~\cite{catseg} learn to convert image-text representations into dense predictions, while TPOV-Seg~\cite{tpovseg} adapts text prompts to remote-sensing segmentation. Training-free methods use fixed pretrained models. MaskCLIP~\cite{maskclip} extracts dense predictions from CLIP, and SCLIP~\cite{sclip} and NACLIP~\cite{naclip} improve local correspondence by modifying attention. For remote sensing, SegEarth-OV~\cite{segearth} upsamples patch features and suppresses global feature bias. ReAttnCLIP~\cite{reattn} revises patch interactions and class-to-patch attention. SAM 3~\cite{sam3} directly produces semantic responses, instance masks, and concept-presence scores for text queries. SegEarth-OV3~\cite{segearth3} combines these three outputs to obtain class predictions.

\begin{figure}[!t]
\centering
\includegraphics[width=\columnwidth]{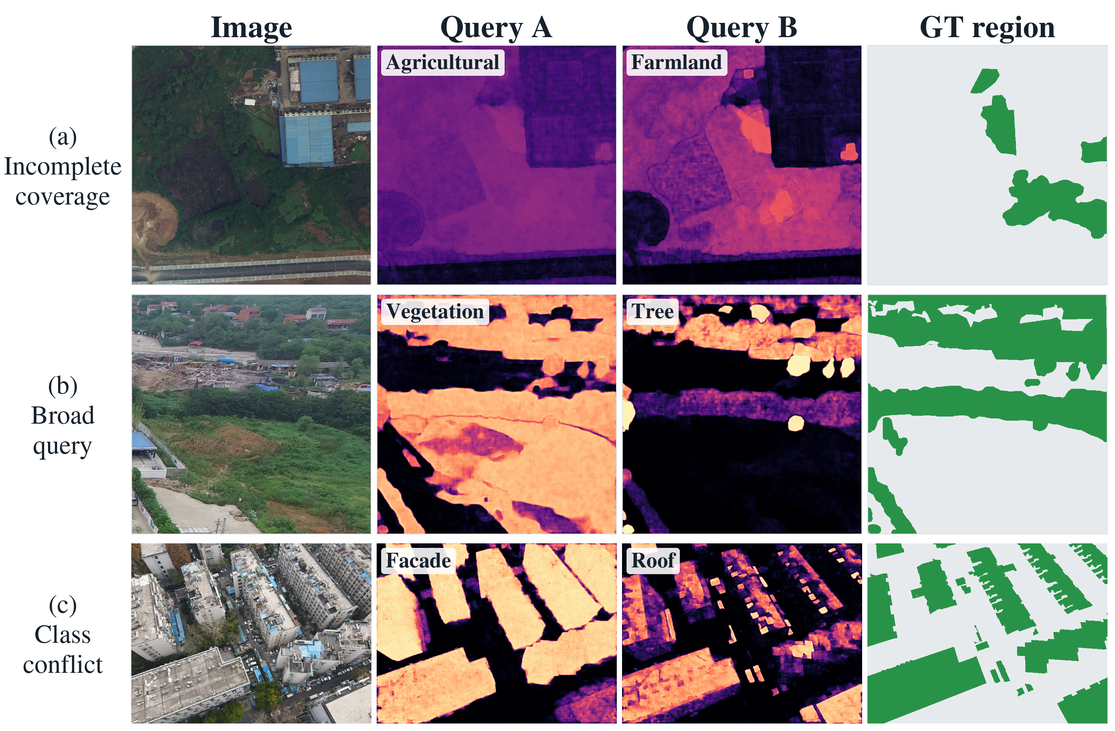}
\caption{SAM 3 responses to three images and their queries. In (a), farmland strengthens responses on regions weakly activated by agricultural. In (b), vegetation activates both low vegetation and trees. In (c), facade responds more strongly than roof on annotated roofs. Heatmaps range from 0 (dark) to 1 (light). Green regions mark agriculture, tree, and roof annotations, respectively.}
\label{fig:motivation}
\end{figure}

These approaches make text-conditioned responses a direct part of pixel-level classification. For the same class, different text phrases can substantially affect SAM 3's responses. A narrow phrase may respond to only part of its intended class, while a broad phrase may also activate regions belonging to other classes. Figure~\ref{fig:motivation} illustrates three cases of SAM 3's text-conditioned responses. In Fig.~\ref{fig:motivation}(a), \emph{agricultural} responds weakly on parts of the annotated farmland, whereas \emph{farmland} gives stronger responses in those regions. In Fig.~\ref{fig:motivation}(b), \emph{vegetation} activates both low vegetation and trees. Assigning this phrase to low vegetation brings the tree responses into that class. In Fig.~\ref{fig:motivation}(c), \emph{facade} produces higher responses than \emph{roof} on annotated roof regions. When both phrases are queried together, many roof pixels in this example are assigned to \emph{facade}. The examples show incomplete coverage and cross-class responses, together with the effect of relative response strength on pixel labels.

Phrase expansion addresses incomplete coverage by allowing several queries to describe the same class. SegEarth-OV uses multiple queries for some classes, such as \emph{building}, \emph{house}, and \emph{roof}, and takes their maximum pixel probability. SegEarth-OV3 also supplements some class names with synonyms and uses the highest probability among the prompts for each class.

However, direct phrase expansion still has limitations. A new phrase may activate regions missed by the initial query or repeat responses already supplied by other phrases. Broad phrases may also produce strong responses on regions belonging to other classes, affecting the final segmentation labels of those pixels. For example, the broad candidate phrase \emph{vegetation} in Fig.~\ref{fig:motivation}(b) activates tree crowns as well as low vegetation. Including it in the low-vegetation phrase set can assign tree pixels to that class. The foreground phrase set therefore needs to be screened to select phrases that provide complementary class coverage and distinguish their assigned class from competing classes.

Semantically related classes can also produce overlapping responses on the same regions. Their response strengths may differ, allowing one class to dominate the pixel-level comparison and absorb regions belonging to the other. This creates ambiguity when the target labels require these related concepts to be distinguished. In Fig.~\ref{fig:motivation}(c), for example, the stronger \emph{facade} response causes roof pixels to be assigned to \emph{facade} when \emph{facade} and \emph{roof} are queried together.

To address phrase selection and inter-class conflicts, we propose PhraseSAM, a training-free framework built on frozen SAM 3, with phrase expansion and selection followed by directed correction. In the first stage, an LLM generates foreground candidate phrases and a background phrase set from the full class list and class definitions. For each foreground candidate phrase, we average responses on reference cores in the unlabeled image pool to measure support for its assigned class and separation from competing classes. We also measure its unique coverage over other candidate phrases in the same class across the full configuration grids. The retained foreground phrases and all generated background phrases form the selected phrase set.

In the second stage, the responses on reference cores identify one competing class for each eligible foreground class. The log-score differences on calibration pixels determine a correction threshold for each proposed correction direction. At inference, the initial label identifies the proposed target, and the correction threshold helps determine whether to update the label.

\begin{itemize}
\item We introduce phrase selection using phrase responses on reference cores and unique coverage. The procedure selects complementary foreground phrases, measures their separation from competing classes, and incorporates a complete background phrase set according to the output-label definitions.
\item We use mutual responses on reference cores to select competing classes and calibration pixels from the same image pool to determine correction thresholds.
\item PhraseSAM achieves 60.5\% mean mIoU on eight remote-sensing datasets, 6.3 percentage points above GLoP-SAM3 (54.2\%). It obtains the highest training-free mIoU among the compared methods on seven datasets.
\end{itemize}

\section{Related Work}

\subsection{Vision-Language and Visual Foundation Models}
Vision-language models connect visual representations with natural language. CLIP~\cite{clip} learns a shared image-text embedding space through contrastive training, allowing category descriptions to serve as recognition queries. Generative models further connect image features with a large language model (LLM) so that visual content can be described and discussed. BLIP-2~\cite{blip2} uses a trainable Querying Transformer between frozen image and language models. It first learns text-relevant visual features and then aligns those features with language generation. LLaVA~\cite{llava} projects vision-encoder features into the language model's input space and uses visual instruction tuning to develop multimodal dialogue. Conversations and image-based questions teach the model to produce responses conditioned on both an image and a user instruction.

Visual foundation models also provide reusable spatial representations for these systems. DINOv2~\cite{dinov2} combines image-level self-distillation and patch-level learning on a large curated image collection. Its image and patch features support recognition and dense prediction, including spatial matching and segmentation. Promptable segmentation models represent regions explicitly as masks. SAM~\cite{sam} combines an image encoder, a prompt encoder, and a mask decoder, accepting spatial prompts such as points, boxes, and masks. This design allows the same pretrained model to delineate different objects specified by the user. SAM 2~\cite{sam2} extends promptable segmentation to images and videos through a streaming memory architecture. Stored information from earlier frames guides the segmentation of the prompted object in subsequent frames.

Language-conditioned detection provides one route from text to object masks. Grounding DINO~\cite{groundingdino} localizes objects described by class names or referring expressions. Grounded SAM~\cite{groundedsam} passes the resulting boxes to SAM to obtain masks. SAM 3 performs detection, segmentation, and tracking from concept prompts specified by a short noun phrase, an image exemplar, or their combination. A single query can return masks for multiple matching instances. The image detector and memory-based video tracker share a backbone, and a presence head estimates whether the concept appears in the image. For semantic segmentation, the semantic response covers regions, the instance masks delineate individual objects, and the presence score indicates support for the query. The resulting pixel labels therefore depend on how the query describes the intended category.

Remote-sensing imagery places additional demands on visual pretraining because the same area can be observed by different sensors and at different dates. CROMA~\cite{croma} uses paired optical and radar observations to align sensor-specific features through contrastive learning. It also reconstructs masked patches from fused features. SkySense~\cite{skysensebase} learns from optical and synthetic aperture radar (SAR) time series. Its encoder first extracts spatial features for individual observations and then combines them across sensors and dates. The resulting features support tasks ranging from scene classification to pixel-level mapping.

Text-based interpretation also requires aligning remote-sensing features with descriptions of objects and land cover. RemoteCLIP~\cite{remoteclip} converts detection and segmentation annotations into image-text pairs. RS5M~\cite{georsclip} assembles a large corpus of remote-sensing images and captions to train GeoRSCLIP. Zavras et al.~\cite{modalitygap} focus on reducing the gap between visual and textual representations through cross-modal alignment. TimeSenCLIP~\cite{timesenclip} extends image-text alignment to sequences of repeated observations.

Building on image-text alignment, instruction tuning trains remote-sensing models to answer questions about scenes, describe specified regions, and return object locations. RSGPT~\cite{rsgpt} fine-tunes InstructBLIP on RSICap, whose captions describe aerial scenes and record object attributes such as shape, position, and count. GeoChat~\cite{geochat} fine-tunes LLaVA-1.5 on scene questions, region descriptions, and grounded dialogue. Its bounding-box format lets users specify a region to describe and lets the model return the location of a referred object. LHRS-Bot~\cite{lhrsbot} builds its alignment and instruction datasets using volunteered geographic information. It connects visual features from several encoder levels to the language model and uses curriculum learning to introduce more complex tasks. SkyEyeGPT~\cite{skyeyegpt} uses a shared instruction format for image description, phrase grounding, visual question answering, and UAV video captioning. Its two-stage tuning covers individual tasks and multi-turn dialogue.

Remote-sensing instruction datasets also bring observations from different sensors and dates into these tasks. In MMRS-1M, EarthGPT~\cite{earthgpt} pairs optical, SAR, and infrared images with instructions for scene description, visual question answering, and object grounding. Its visual encoder combines transformer and convolutional features to capture scene context and local detail. EarthDial~\cite{earthdial} includes multispectral and radar observations at multiple resolutions. It also uses temporal sequences to train the model to compare observations of the same location, as required in change detection.

\subsection{Open-Vocabulary Semantic Segmentation}
Open-vocabulary semantic segmentation labels image pixels using text-defined categories, including categories without pixel-level training annotations. Trained open-vocabulary semantic segmentation methods learn to convert image-text representations into spatial class predictions. SAN predicts masks and classifies them with CLIP. SCAN~\cite{scan} calibrates region representations using CLIP's semantic knowledge. CAT-Seg learns to aggregate dense image-text matching costs, while SED~\cite{sed} combines features from a hierarchical encoder in its decoder. For remote sensing, OVRS~\cite{ovrs} addresses varying object orientations and scales through rotation-aware matching and multiscale feature integration. ROSS~\cite{ross} combines rotation-aware features with features from a remote-sensing encoder. TPOV-Seg adapts text prompts to land-cover semantics, and Zermatten et al.~\cite{landsemantics} align pixel features with land-cover descriptions through supervised contrastive learning. Knowledge graph-enhanced class representations~\cite{knowledgegraph} incorporate relations such as synonyms and class hierarchies. Soft-Guided OVSS~\cite{softguided} uses soft guidance to adapt a vision-language model, and text-to-image activation~\cite{textactivation} produces spatial evidence conditioned on the target class. Joint modeling of semantic and structural cues~\cite{rsovsg} and refinement with remote-sensing priors~\cite{cdseg} address errors in category predictions.

Training-free CLIP methods obtain dense predictions from frozen image-text representations. MaskCLIP establishes matching between pixels and text using spatial visual features. SCLIP uses self-self attention to improve local correspondence. GEM~\cite{gem} uses self-self attention and groups tokens iteratively, while NACLIP weights interactions by spatial neighborhood. ClearCLIP~\cite{clearclip} decomposes CLIP representations to isolate the local features used for dense prediction. CLIP Surgery~\cite{clipsurgery} suppresses redundant activations. ProxyCLIP~\cite{proxyclip} builds attention from auxiliary visual features, while CorrCLIP~\cite{corrclip} constrains patch correlations using segmentation regions and visual similarity. For remote-sensing images, SegEarth-OV combines higher-resolution features from SimFeatUp with global-bias suppression. ReAttnCLIP adjusts patch similarity and class-to-patch attention. Ye et al.~\cite{costless} also use pretrained foundation models for training-free remote-sensing segmentation.

Open-vocabulary methods based on promptable segmentation models organize concept-conditioned outputs across descriptions and classes. CoCo-SAM3~\cite{coco} aligns and aggregates evidence from synonymous prompts before comparing classes on a common scale. In remote sensing, SegEarth-OV3 combines semantic responses for continuous regions with instance masks for discrete objects, weighting the fused predictions by concept presence. GLoP-SAM3~\cite{glopsam3} incorporates global and local priors through reliability gating.

The query list determines which responses contribute to each output class. SegEarth-OV adjusts the queries to the dataset categories. For example, it uses \emph{building}, \emph{house}, and \emph{roof} for the building class and takes the highest pixel probability across these queries. SegEarth-OV3 starts from class names, adds synonyms to some classes, and pools their predictions by a pixel-wise maximum. PhraseSAM evaluates candidate phrase responses on the image pool to select complementary foreground phrases and configure corrections between competing classes.

\section{Method}

\begin{figure*}[!tbp]
\centering
\includegraphics[width=\textwidth]{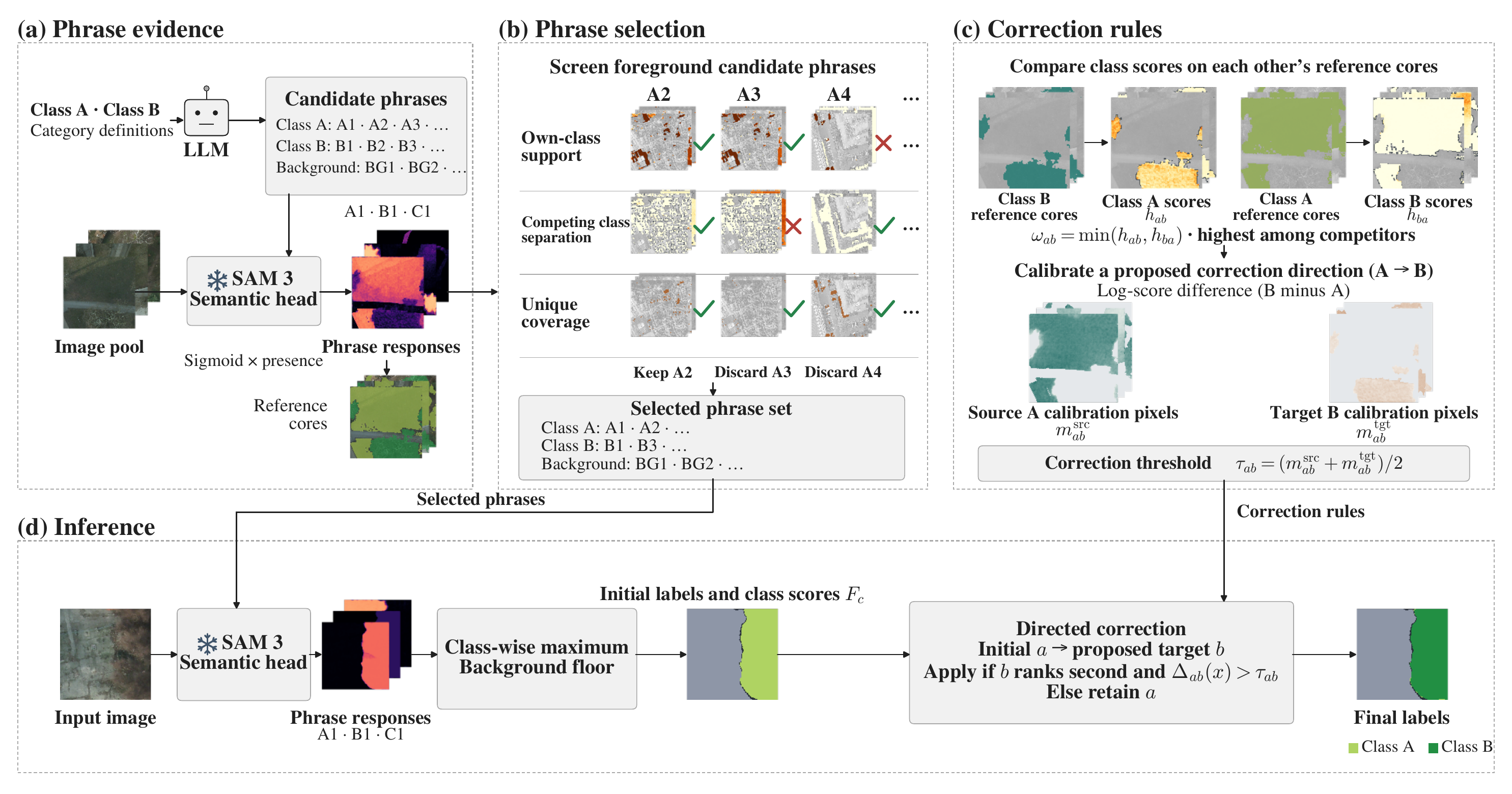}
\caption{PhraseSAM uses the image pool to select phrases and configure correction rules. In (a), the initial queries define the reference cores. Class and phrase identifiers are illustrative. In (b), checks and crosses show the screening decisions for each example in its image pool. A2 and A3 use OpenEarthMap, and A4 uses VDD. The selected phrase set combines retained foreground phrases with all generated background phrases. In (c), the upper stacks show how each class responds on the other class's reference cores. The lower maps show log-score differences on the calibration pixels. Teal and copper indicate negative and positive differences. The midpoint of the source and target group medians defines the correction threshold. Panel (d) shows the resulting Class A-to-B correction on a Potsdam crop.}
\label{fig:overview}
\end{figure*}

Open-vocabulary segmentation starts with a list of target classes. PhraseSAM uses the full class list and an unlabeled RGB image pool (Fig.~\ref{fig:overview}). Each class has an initial query. First, an LLM generates foreground candidate phrases and background phrases. We use frozen SAM 3 responses on the image pool to screen the foreground candidate phrases. The generated background phrases are used in full. Together with the retained foreground phrases, they form the selected phrase set. Second, we compute class scores from these phrases on the same image pool to select competing classes and calibrate correction rules.

At inference, frozen SAM 3 produces a response map for each selected phrase for the new RGB image. We take the maximum phrase response within each class to compute the class scores and assign the initial labels. Each pixel uses its initial label to look up a correction rule. The rule compares the source and target class scores to decide whether to update the label.

Let $K$ be the number of output classes and $\mathcal C=\{0,\ldots,K-1\}$ their index set. The background index $b_0\in\mathcal C$ denotes the residual label, such as Clutter in Potsdam and Vaihingen. For class $c$, let $w_c^0$ denote the initial query, $\mathcal W_c$ the candidate phrase set, and $\mathcal A_c$ the selected phrase set. A candidate phrase is written as $w\in\mathcal W_c$. The image pool $\mathcal I=\{I_i\}_{i=1}^{N}$ contains $N$ RGB images, with $I_i$ denoting the $i$th image.

\begin{figure}[!t]
\centering
\includegraphics[width=\columnwidth]{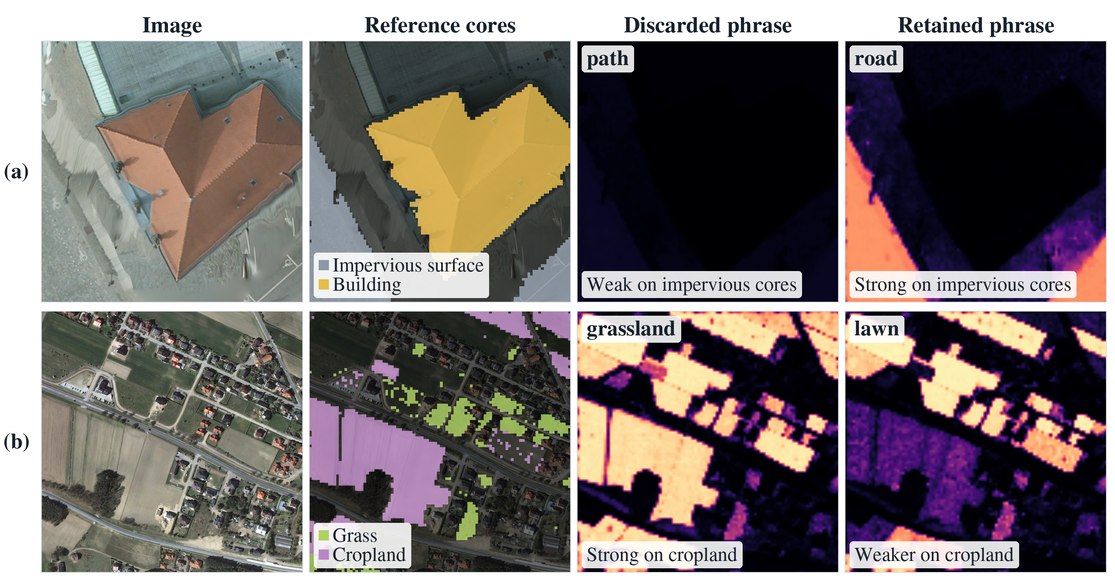}
\caption{Reference cores and phrase responses on Potsdam (a) and OpenEarthMap (b). All heatmaps use a common scale from 0 to 1.}
\label{fig:anchor-cores}
\end{figure}

\subsection{Phrase Expansion and Selection}
\label{sec:phrase-expansion}
\label{sec:semantic-configuration}
\subsubsection{Foreground and Background Phrases}
Given a target dataset, we begin with phrase generation. We provide the LLM with the full class list, with a definition and an initial query for each class. For each foreground class, the LLM generates short phrases that name the target object or region or describe how it typically looks in aerial images. The candidate phrase set $\mathcal W_c$ contains these generated phrases and the initial query $w_c^0$. Category qualifiers are preserved when they distinguish output labels, such as \emph{large} and \emph{small} for vehicles. The initial query provides the starting phrase and may differ from the official label name: for example, \emph{road} initializes the impervious-surface class in Potsdam.

The background phrase set describes the content assigned to the residual label. The LLM generates it directly from the full class list and its definitions, so its scope depends on which categories the dataset treats as foreground. For example, \emph{open water} belongs to the background phrase set in UDD5, whereas VDD has a foreground water class. Likewise, \emph{roof} is a background phrase in iSAID and a foreground class in VDD. All generated background phrases form $\mathcal W_{b_0}$ and are used directly to compute the background score. This set stays fixed during foreground phrase selection. Table~\ref{tab:inventory-all} provides the candidate and selected phrases for every output class across all eight datasets.

\subsubsection{SAM 3 Phrase Responses and Reference Cores}
We evaluate the candidate phrases on the image pool. Given image $I_i$ and phrase $w$, SAM 3 outputs semantic logits $L_{i,w}$ and concept-presence probability $p_{i,w}$. For phrase selection, we sample the native logits with stride $s_g$ along both axes to obtain the configuration grid $\mathcal G_i$. At a zero-based grid location $z$, the phrase response is given by
\begin{equation}
 \widetilde R_{i,w}(z)=p_{i,w}\,\sigma\!\left(L_{i,w}(s_g z)\right),
 \label{eq:grid-response}
\end{equation}
where $\sigma$ is the sigmoid function.

The initial queries $w_c^0$ of all output classes, including background, define the reference cores used for phrase selection and for selecting competing classes. At each grid location, the initial-query responses compete across classes. A location belongs to the reference core $\mathcal H_{i,c}$ when class $c$ has the largest response, its response reaches the fixed minimum $\tau_{\mathrm{core}}$, and it is at least $\rho_{\mathrm{core}}$ times the second-highest response, where $\rho_{\mathrm{core}}$ is a fixed ratio. Counting the reference locations across the image pool gives $n_c=\sum_i|\mathcal H_{i,c}|$ for class $c$. We require at least $n_{\mathrm{core}}$ reference locations before using a class in response comparisons. With very few locations, the response in a small region can dominate the mean. Figure~\ref{fig:anchor-cores} shows examples of reference cores and phrase responses measured on them.

\subsubsection{Class Support and Unique Coverage}
\label{sec:phrase-roles}
\label{sec:selection-details}
Let $\mathcal C_{\mathrm{rel}}=\{c\in\mathcal C:n_c\geq n_{\mathrm{core}}\}$ denote the classes that meet the minimum. For each foreground candidate phrase $w\in\mathcal W_c$, we compare its mean response on its assigned class's reference cores with the largest mean on the reference cores of competing classes in $\mathcal C_{\mathrm{rel}}$. The own-class mean, highest competing-class mean, and relative separation are defined as
\begin{equation}
\begin{aligned}
 \mu_{w,c}&=\frac{1}{n_c}\sum_{i=1}^{N}\sum_{z\in\mathcal H_{i,c}}\widetilde R_{i,w}(z),\\
 \mu^{\mathrm{rival}}_{w,c}&=\max_{\substack{d\in\mathcal C_{\mathrm{rel}}\\d\ne c}}\mu_{w,d},\\
 q_{w,c}&=\frac{\mu_{w,c}-\mu^{\mathrm{rival}}_{w,c}}{\mu_{w,c}}.
\end{aligned}
 \label{eq:specificity}
\end{equation}
The mean $\mu_{w,c}$ uses all reference core locations of class $c$ in the image pool. The sampled responses $\widetilde R_{i,w}(z)$ lie in $[0,1]$ by Eq.~\eqref{eq:grid-response}, so $\mu_{w,c}$ is nonnegative. We compute $q_{w,c}$ only when $\mu_{w,c}>0$. When $\mu_{w,c}=0$, relative separation is undefined, and a generated candidate phrase fails the positive own-class threshold in Eq.~\eqref{eq:phrase-admission}.

The response comparisons on reference cores do not measure whether a phrase adds spatial coverage to its class phrase set. We assess that contribution using the full configuration grids. Let $V_{i,c}(z)=\max_{v\in\mathcal W_c}\widetilde R_{i,v}(z)$ be the largest response among the candidate phrases in class $c$. A grid location counts toward the unique coverage of phrase $w$ when
\begin{equation}
\begin{aligned}
 c&=\arg\max_{d\in\mathcal C}V_{i,d}(z),\\
 \widetilde R_{i,w}(z)&>\max_{v\in\mathcal W_c\setminus\{w\}}\widetilde R_{i,v}(z)+\epsilon_u.
\end{aligned}
 \label{eq:unique-support}
\end{equation}
The first condition compares the maximum candidate phrase responses across classes. If several classes share the maximum, the class with the smallest index wins. The second condition requires $w$ to outscore every other candidate phrase in its class by more than the fixed margin $\epsilon_u$. For a class with a single candidate phrase, the comparison response on the right-hand side is set to zero before adding $\epsilon_u$. The unique coverage count $u_w$ is the number of locations satisfying both conditions across the full configuration grids of the image pool $\mathcal I$.

Each foreground phrase set starts from its initial query $w_c^0$. We screen generated foreground phrases only if $\mathcal C_{\mathrm{rel}}$ contains both the assigned class and at least one competing class. Otherwise, the phrase set retains its initial query. A generated foreground phrase passes screening when
\begin{equation}
 \mu_{w,c}\geq\tau_{\mathrm{own}},\qquad
 \mu_{w,c}\geq\rho_{\mathrm{sep}}\mu^{\mathrm{rival}}_{w,c},\qquad
 u_w>0.
 \label{eq:phrase-admission}
\end{equation}
The fixed threshold $\tau_{\mathrm{own}}$ sets the minimum mean response on the assigned class's reference core. The fixed ratio $\rho_{\mathrm{sep}}$ sets how large this mean must be relative to the highest competing-class mean. The last condition requires at least one grid location of unique coverage.

If the highest competing-class mean of the initial query $w_c^0$ exceeds the fixed threshold $\tau_{\mathrm{cross}}$, we look for a replacement among the other candidate phrases in $\mathcal W_c$. For each such phrase $w$ that passes Eq.~\eqref{eq:phrase-admission}, we compute $q_{w,c}$ from the responses to $w$ on the reference cores of its assigned class and competing classes. The candidate phrase qualifies as a replacement if this score reaches the fixed threshold $\eta_{\mathrm{sep}}$.
\begin{equation}
 \underbrace{\mu^{\mathrm{rival}}_{w_c^0,c}>\tau_{\mathrm{cross}}}_{\text{initial-query check}},
 \qquad
 \underbrace{q_{w,c}\geq\eta_{\mathrm{sep}}}_{\text{replacement separation}}.
 \label{eq:initial-replacement}
\end{equation}
If the check is triggered and at least one alternative qualifies, we retain all qualifying replacement phrases. The initial query is removed only when both of these conditions hold and it differs from the official class name.

Foreground selection yields the selected phrase set $\mathcal A_c$ for each foreground class. The selected background phrase set is $\mathcal A_{b_0}=\mathcal W_{b_0}$.

\subsubsection{From Phrase Responses to Class Labels}
\label{sec:initial-readout}
On the configuration grid, the class score is $\widetilde F_{i,c}(z)=\max_{w\in\mathcal A_c}\widetilde R_{i,w}(z)$. These grid class scores are used to select competing classes.

Threshold calibration and inference use class scores at the input image resolution. Let $\mathcal U$ denote bilinear interpolation of the native logits to the input image dimensions. Omitting the image index, we compute the phrase response $R_w(x)$ and the class score $F_c(x)$ at pixel $x$ as
\begin{equation}
 \begin{aligned}
 R_w(x)&=p_w\,\sigma\!\left(\mathcal U(L_w)(x)\right),\\
 F_c(x)&=\max_{w\in\mathcal A_c}R_w(x).
 \end{aligned}
 \label{eq:classscore}
\end{equation}
We compute grid and input-resolution class scores for all classes, including background.

Before prediction labels are assigned, the background score is raised to $\max(F_{b_0}(x),\beta_{b_0})$, where $\beta_{b_0}$ is the dataset-specific background floor. The highest score after applying the background floor determines the initial label $y_0(x)$, with ties resolved in class-index order. Section~\ref{sec:implementation} describes the background settings. Directed correction takes $y_0$ and the unadjusted class scores $F_c$ as inputs.

\subsection{Directed Correction}
\label{sec:residual-conflict-correction}
\subsubsection{Selecting a Competing Class}
\label{sec:conflict-discovery}
\label{sec:direction-correction}
Two foreground classes can still respond to the same region after phrase selection. For example, a strong facade response can cause a roof pixel to receive the facade label. We use the class scores on the image pool to propose correction directions and calibrate their thresholds.

We consider a foreground source class $a$ and a distinct foreground class $b$ if each has at least $n_{\mathrm{core}}$ reference locations across the image pool. Let $h_{ab}$ be the average grid score of class $a$ on the $n_b$ reference locations of class $b$. Exchanging $a$ and $b$ gives $h_{ba}$. We choose the proposed target $t(a)$ by maximizing the smaller of these two means:
\begin{equation}
 \begin{aligned}
 h_{ab}&=\frac{1}{n_b}\sum_{i=1}^{N}\sum_{z\in\mathcal H_{i,b}}\widetilde F_{i,a}(z),\\
 \omega_{ab}&=\min(h_{ab},h_{ba}),\quad
 t(a)=\arg\max_{b\in\mathcal C_{\mathrm{rel}}\setminus\{b_0,a\}}\omega_{ab}.
 \end{aligned}
 \label{eq:competition}
\end{equation}
The minimum $\omega_{ab}$ measures how strongly the two classes respond on each other's reference cores. Each source class $a$ independently proposes one correction direction $(a,t(a))$. We set $t(c)=c$ for background and foreground classes that lack sufficient reference locations or an eligible competing class.

\subsubsection{Calibrating Correction Thresholds}
\label{sec:bidirectional-competition}
\label{sec:correction-configuration}
To calibrate $a\to b$, we sample the input-resolution class scores in each image at stride $s_{\mathrm{cal}}$ along both axes, before applying the background floor. A sampled pixel is a calibration pixel if its highest score among all classes reaches $\tau_{\mathrm{core}}$ and is at least $\rho_{\mathrm{core}}$ times the second-highest score. Among the calibration pixels, those with $a$ as the top-scoring class form the \emph{source group}, and those with $b$ as the top-scoring class form the \emph{target group}. These groups use the same confidence conditions as the reference cores (Section~\ref{sec:implementation}).

For each calibration pixel $x$ in either group, we subtract the log score of $a$ from that of $b$.
\begin{equation}
 \begin{aligned}
 \Delta_{ab}(x)={}&\log\max(F_b(x),\epsilon_\ell)\\
 &-\log\max(F_a(x),\epsilon_\ell),
 \end{aligned}
 \label{eq:raw-margin}
\end{equation}
The positive log-score floor $\epsilon_\ell$ keeps the logarithms defined.

For each group, we take the median of $\Delta_{ab}(x)$ in every image with at least $n_{\mathrm{cal}}$ calibration pixels in that group. We then take a second median across the images that contribute to each group, giving the source group median $m_{ab}^{\mathrm{src}}$ and the target group median $m_{ab}^{\mathrm{tgt}}$. If both groups contribute image-level medians, we take the midpoint of their pooled medians as the correction threshold for $a\to b$.
\begin{equation}
 \tau_{ab}=\frac{m_{ab}^{\mathrm{src}}+m_{ab}^{\mathrm{tgt}}}{2}.
 \label{eq:calibration}
\end{equation}
Let $\mathcal D$ contain the ordered pairs whose estimated thresholds satisfy $\tau_{ab}<0$. An initial foreground label $a$ satisfies $F_a(x)\geq F_b(x)$ for every competing foreground class $b$, so $\Delta_{ab}(x)\leq0$. A correction can therefore occur only for a negative threshold.

\subsubsection{Applying Correction Rules at Inference}
For each pixel $x$ in a new image, the class with the highest score after applying the background floor gives the initial label $y_0(x)$ (Section~\ref{sec:initial-readout}). Let $a=y_0(x)$ and $b=t(a)$ be its proposed target. Among the remaining classes, let $r(x)=\arg\max_{c\in\mathcal C\setminus\{a\}}F_c(x)$ be the class with the highest unadjusted score, using the same tie convention. A correction requires $b=r(x)$ and a log-score difference that exceeds the correction threshold. We compute $\Delta_{ab}(x)$ from the unadjusted class scores. The final label is
\begin{equation}
 \widehat y(x)=
 \begin{cases}
 b, & (a,b)\in\mathcal D,\ b=r(x),\ \Delta_{ab}(x)>\tau_{ab},\\
 a, & \text{otherwise}.
 \end{cases}
 \label{eq:correction}
\end{equation}
For example, the facade-to-roof rule tests a pixel when facade ranks first and roof ranks second. The negative correction threshold determines how far the roof score can be below the facade score for the pixel to change label. All decisions use the initial label map $y_0$, so each pixel can change label at most once.

\section{Experiments}
\subsection{Evaluation Setup}
We evaluate on LoveDA~\cite{loveda}, Potsdam and Vaihingen~\cite{isprs2d}, iSAID~\cite{isaid}, OpenEarthMap~\cite{openearthmap}, UDD5~\cite{udd5}, VDD~\cite{vdd}, and UAVid~\cite{uavid}. We refer to these eight datasets collectively as OVR-S8. We follow SegEarth-OV3 for class definitions, initial queries, and label mappings. The collection covers land-cover regions, urban surfaces, and aerial objects. Within each dataset, we sum native-pixel confusion matrices and average IoU over the output labels, including background. The eight-dataset mean weights datasets equally. Scores use one decimal place; differences are computed before rounding.

In practice, training-free open-vocabulary segmentation uses a single unlabeled collection, with no training, validation, or test partition. On OVR-S8, scenes and class distributions may differ across splits, making estimates from training images less representative of the test images. Accordingly, the main experiments sample the image pool directly from the unlabeled test images to select phrases and configure correction rules for that same collection (Table~\ref{tab:data-protocol}).

Nevertheless, Table~\ref{tab:main-results} also reports results with a separate image pool. We randomly sample from the official training splits of seven datasets and use the official validation split for VDD. Both settings use the same test images.

\begin{table*}[!t]
\centering
\small
\renewcommand{\arraystretch}{.86}
\setlength{\tabcolsep}{4pt}
\caption{Dataset sizes and phrase settings. Counts of candidate and selected phrases include background phrases; BG gives the fixed background phrase count.}
\label{tab:data-protocol}
\begin{tabular*}{\textwidth}{@{\extracolsep{\fill}}lrrrrrr@{}}
\toprule
Dataset & Inputs & Classes & Image pool size & Candidate phrases & Selected phrases & BG\\
\midrule
LoveDA & 1,669 & 7 & 256 & 41 & 27 & 6\\
Potsdam & 2,016 & 6 & 256 & 38 & 29 & 9\\
Vaihingen & 398 & 6 & 256 & 37 & 29 & 8\\
iSAID & 11,644 & 16 & 256 & 71 & 59 & 18\\
OpenEarthMap & 384 & 9 & 256 & 51 & 39 & 7\\
UDD5 & 40 & 5 & 40 & 34 & 30 & 13\\
VDD & 40 & 7 & 40 & 42 & 33 & 10\\
UAVid & 560 & 7 & 256 & 56 & 50 & 23\\
\bottomrule
\end{tabular*}
\par\vspace{8pt}
{
\centering
\setlength{\aboverulesep}{0.25ex}
\setlength{\belowrulesep}{0.25ex}
\caption{OVR-S8 results in mIoU (\%). Bold and underlining mark the best and second-best training-free scores in the main comparison. Mean weights all eight datasets equally.}
\label{tab:main-results}
\small
\setlength{\tabcolsep}{3pt}
\renewcommand{\arraystretch}{1.04}
\begin{tabular*}{\textwidth}{@{\extracolsep{\fill}}lrrrrrrrrr@{}}
\toprule
Method & LoveDA & Potsdam & Vaihingen & iSAID & OpenEarthMap & UDD5 & VDD & UAVid & Mean\\
\midrule
\multicolumn{10}{l}{\color[gray]{0.42}\textit{Trained, closed-set}}\\
\textcolor[gray]{0.42}{SegFormer-b0~\cite{segformer}} & \textcolor[gray]{0.42}{50.0} & \textcolor[gray]{0.42}{74.3} & \textcolor[gray]{0.42}{61.2} & \textcolor[gray]{0.42}{36.2} & \textcolor[gray]{0.42}{64.4} & \textcolor[gray]{0.42}{56.5} & \textcolor[gray]{0.42}{62.9} & \textcolor[gray]{0.42}{59.7} & \textcolor[gray]{0.42}{58.2}\\
\midrule
\multicolumn{10}{l}{\textit{Trained, open-vocabulary}}\\
SCAN~\cite{scan} & 23.2 & 27.5 & 15.2 & 44.3 & --- & 34.1 & 29.2 & 20.3 & ---\\
SAN~\cite{san} & 25.3 & 37.3 & 39.2 & 49.6 & --- & 37.2 & 35.8 & 23.5 & ---\\
SED~\cite{sed} & 24.6 & 29.4 & 39.0 & 51.2 & --- & 35.7 & 32.5 & 21.3 & ---\\
CAT-Seg~\cite{catseg} & 28.6 & 35.8 & 42.3 & 53.3 & --- & 40.2 & 39.1 & 25.7 & ---\\
OVRS~\cite{ovrs} & 31.5 & 36.4 & 43.5 & 52.7 & --- & 40.8 & 37.2 & 24.1 & ---\\
GSNet~\cite{gsnet} & 32.5 & 37.9 & 44.1 & 53.7 & --- & 40.9 & 37.3 & 24.2 & ---\\
RSKT-Seg~\cite{rskt} & 33.2 & 38.4 & 42.7 & 54.3 & --- & 42.1 & 39.7 & 25.7 & ---\\
SkySense-O~\cite{skysense} & 38.3 & 54.1 & 51.6 & 43.9 & 40.8 & --- & --- & --- & ---\\
\midrule
\multicolumn{10}{l}{\textit{Training-free, open-vocabulary}}\\
CLIP~\cite{clip} & 12.4 & 15.6 & 10.8 & 7.5 & 12.0 & 9.5 & 14.2 & 10.9 & 11.6\\
MaskCLIP~\cite{maskclip} & 27.8 & 33.9 & 29.9 & 14.5 & 25.1 & 32.4 & 32.9 & 28.6 & 28.1\\
SCLIP~\cite{sclip} & 30.4 & 39.6 & 35.9 & 16.1 & 29.3 & 38.7 & 37.9 & 31.4 & 32.4\\
GEM~\cite{gem} & 31.6 & 39.1 & 36.4 & 17.7 & 33.9 & 41.2 & 39.5 & 33.4 & 34.1\\
ClearCLIP~\cite{clearclip} & 32.4 & 42.0 & 36.2 & 18.2 & 31.0 & 41.8 & 39.3 & 36.2 & 34.6\\
SegEarth-OV~\cite{segearth} & 36.9 & 48.5 & 40.0 & 21.7 & 40.3 & 50.6 & 45.3 & 42.5 & 40.7\\
ProxyCLIP~\cite{proxyclip} & 34.3 & 49.0 & 47.5 & 21.8 & 38.9 & 40.8 & 47.8 & 35.8 & 39.5\\
CorrCLIP~\cite{corrclip} & 36.9 & 51.9 & 47.0 & 25.5 & 32.9 & 46.1 & 47.3 & 38.3 & 40.7\\
SegEarth-OV3~\cite{segearth3} & \underline{47.4} & 57.8 & 60.8 & 27.6 & 42.9 & 71.7 & 64.5 & 54.7 & 53.4\\
SegEarth-OV3 (no tiling or pre-upsampling) & \underline{47.4} & 57.8 & 60.8 & 27.5 & 42.4 & 71.7 & 64.4 & \underline{60.1} & 54.0\\
GLoP-SAM3~\cite{glopsam3} & 46.2 & \underline{57.9} & \underline{62.2} & \underline{28.0} & \textbf{46.6} & \underline{72.6} & \underline{64.7} & 55.4 & \underline{54.2}\\
\midrule
\textcolor[gray]{0.42}{PhraseSAM (Ours)\textsuperscript{*}} & \textcolor[gray]{0.42}{48.4} & \textcolor[gray]{0.42}{62.4} & \textcolor[gray]{0.42}{67.5} & \textcolor[gray]{0.42}{38.5} & \textcolor[gray]{0.42}{46.5} & \textcolor[gray]{0.42}{74.5} & \textcolor[gray]{0.42}{74.9} & \textcolor[gray]{0.42}{67.3} & \textcolor[gray]{0.42}{60.0}\\
PhraseSAM (Ours) & \textbf{48.4} & \textbf{63.2} & \textbf{67.7} & \textbf{39.7} & \underline{46.4} & \textbf{74.5} & \textbf{75.4} & \textbf{69.0} & \textbf{60.5}\\
\bottomrule
\end{tabular*}
}
\end{table*}

\subsection{Implementation}
\label{sec:implementation}
We use frozen SAM 3 with processor resolution 1008 and FP16 inference on NVIDIA V100 GPUs. The candidate phrase set includes the initial queries. The shared instruction requests whole-object or region phrases, class coverage, preserved label qualifiers, and background phrases. It restricts traffic-road synonyms to roadway phrases. The candidate and selected phrase sets are listed in Table~\ref{tab:inventory-all}.

Each image pool contains up to 256 images. Following SegEarth-OV and SegEarth-OV3, we tune the background floor for each dataset.

The native $288\times288$ semantic logits yield a $96\times96$ configuration grid at stride $s_g=3$. We select both reference cores and calibration pixels with $\tau_{\mathrm{core}}=0.5$ and $\rho_{\mathrm{core}}=1.5$. Class response comparisons require at least $n_{\mathrm{core}}=32$ reference locations. Phrase selection uses $\tau_{\mathrm{own}}=0.2$ and $\rho_{\mathrm{sep}}=1.5$, and the initial-query and replacement tests use $\tau_{\mathrm{cross}}=0.4$ and $\eta_{\mathrm{sep}}=0.8$. Calibration uses $s_{\mathrm{cal}}=8$ and $n_{\mathrm{cal}}=1$ to estimate $\tau_{ab}$ from the image pool. We set $\epsilon_u=10^{-4}$ for unique coverage and $\epsilon_\ell=10^{-4}$ for the log scores. All datasets share these settings.

At inference, each evaluated image or prepared tile is processed at resolution 1008, and class scores are computed at its input resolution using Eq.~\eqref{eq:classscore}.

For the comparison figures, SegEarth-OV3 predictions use its released semantic, instance, and presence aggregation under FP16 autocast. For OpenEarthMap, SegEarth-OV3 uses its released 512/512 sliding-window setting; for the other datasets, it processes complete evaluated inputs. SegEarth-OV and CorrCLIP images come from their saved native-resolution predictions.

\subsection{Comparison With Existing Methods}
Table~\ref{tab:main-results} compares PhraseSAM with eight trained open-vocabulary methods and ten training-free baselines. SegFormer-b0 supplies the trained closed-set reference. Published scores come from Table~1 of SegEarth-OV3 and Table~II of GLoP-SAM3. We also run SegEarth-OV3 on the same evaluation inputs and label mappings as PhraseSAM, using a processor resolution of 1008 with sliding windows disabled. We compute the arithmetic mean over the eight scores for each method with complete results. PhraseSAM uses the prepared inventories and input counts in Table~\ref{tab:data-protocol}.

PhraseSAM achieves the highest mean mIoU among the training-free methods and leads on seven of the eight datasets.

The main setting in Table~\ref{tab:main-results} reflects how training-free open-vocabulary segmentation is used in practice. Users typically provide a collection of unlabeled images to segment, so they need no separate training, validation, and test partitions. PhraseSAM samples images from the collection, uses the model responses to assess phrase coverage and identify competing classes, and applies the selected phrases and correction rules to the full collection. On OVR-S8, the official training and test splits can differ in scene composition and class distribution. Sampling from the target images gives response statistics for the scenes to be segmented. In the main experiments, we sample from the unlabeled test images for configuration and obtain 60.5\% mean mIoU across the eight datasets.

Table~\ref{tab:main-results} also reports a separate image pool in the row marked PhraseSAM (Ours)\textsuperscript{*}. We randomly draw configuration images from the official training splits of seven datasets and use the official validation split for VDD. We select phrases and configure correction rules on these images, then apply them to the same test images as in the main setting. With the separate image pool, PhraseSAM reaches 60.0\% mean mIoU, 0.5 percentage points below the main setting and 5.8 points above GLoP-SAM3 (54.2\%).

\vspace{-3pt}
\begin{table*}[!t]
\centering
\par\vspace{3pt}
\renewcommand{\arraystretch}{1.0}
\caption{Final class IoU (\%) on the eight OVR-S8 datasets.}
\label{tab:complete-classes-a}
\label{tab:complete-classes-b}
\footnotesize
\setlength{\tabcolsep}{2pt}
\begin{tabular}{@{}>{\raggedright\arraybackslash}p{.13\textwidth}
  *{9}{>{\raggedright\arraybackslash}p{\dimexpr(.87\textwidth-18\tabcolsep)/9\relax}}@{}}
\toprule
\textbf{LoveDA} & Background & Building & Road & Water & Barren & Forest & Agricultural\\[1pt]
 & 45.7 & 62.6 & 53.1 & 58.0 & 34.3 & 37.1 & 48.1\\[1pt]
\midrule
\textbf{Potsdam} & Clutter (Background) & Impervious surface & Building & Low vegetation & Tree & Car\\[1pt]
 & 22.4 & 77.4 & 86.5 & 60.7 & 50.0 & 82.0\\[1pt]
\midrule
\textbf{Vaihingen} & Clutter (Background) & Impervious surface & Building & Low vegetation & Tree & Car\\[1pt]
 & 36.9 & 80.0 & 86.8 & 57.0 & 74.2 & 71.5\\[1pt]
\midrule
\textbf{iSAID} & Background & Ship & Storage tank & Baseball diamond & Tennis court & Basketball court & Running track & Bridge\\[1pt]
 & 98.2 & 61.9 & 71.4 & 0.0 & 45.6 & 3.1 & 16.0 & 14.3\\[1pt]
\addlinespace[3pt]
 & Large vehicle & Small vehicle & Helicopter & Swimming pool & Roundabout & Soccer field & Plane & Harbor\\[1pt]
 & 62.1 & 36.2 & 7.3 & 37.4 & 35.0 & 49.4 & 73.8 & 24.1\\[1pt]
\midrule
\textbf{OpenEarthMap} & Background & Bareland & Grass & Pavement & Road & Tree & Water & Cropland & Building\\[1pt]
 & 73.4 & 0.3 & 22.3 & 32.3 & 46.9 & 59.1 & 71.7 & 38.5 & 73.3\\[1pt]
\midrule
\textbf{UDD5} & Background & Vegetation & Building & Road & Vehicle\\[1pt]
 & 52.9 & 87.7 & 91.5 & 67.4 & 73.0\\[1pt]
\midrule
\textbf{VDD} & Background & Facade & Road & Vegetation & Vehicle & Roof & Water\\[1pt]
 & 57.9 & 64.4 & 69.4 & 83.6 & 74.1 & 84.5 & 93.9\\[1pt]
\midrule
\textbf{UAVid} & Background & Building & Road & Car & Tree & Low vegetation & Human\\[1pt]
 & 53.7 & 92.0 & 68.0 & 79.3 & 76.8 & 65.8 & 47.4\\[1pt]
\bottomrule
\end{tabular}
\par\vspace{5pt}
\caption{Cumulative component ablations on VDD and Potsdam in IoU and mIoU (\%). Parentheses show the change from the preceding stage in percentage points, computed before rounding.}
\label{tab:additional-component-ablations}
{
\footnotesize
\setlength{\tabcolsep}{2pt}
\begin{tabular*}{\textwidth}{@{\extracolsep{\fill}}l*{8}{l}@{}}
\toprule
\textbf{VDD} & mIoU & Background & Facade & Road & Vegetation & Vehicle & Roof & Water\\
\midrule
Direct & 63.9 & 52.7 & 35.4 & 64.6 & 83.3 & 74.0 & 44.4 & 92.9\\
+ Foreground expansion & 64.9\,{\scriptsize ($+1.0$)} & 42.3\,{\scriptsize ($-10.4$)} & 42.3\,{\scriptsize ($+6.9$)} & 57.8\,{\scriptsize ($-6.8$)} & 82.2\,{\scriptsize ($-1.2$)} & 73.6\,{\scriptsize ($-0.4$)} & 62.3\,{\scriptsize ($+17.9$)} & 94.0\,{\scriptsize ($+1.1$)}\\
+ Background expansion & 66.9\,{\scriptsize ($+1.9$)} & 49.9\,{\scriptsize ($+7.6$)} & 43.2\,{\scriptsize ($+0.9$)} & 61.9\,{\scriptsize ($+4.1$)} & 82.6\,{\scriptsize ($+0.4$)} & 73.7\,{\scriptsize ($+0.2$)} & 62.5\,{\scriptsize ($+0.2$)} & 94.1\,{\scriptsize ($+0.1$)}\\
+ Foreground selection & 69.3\,{\scriptsize ($+2.4$)} & 57.9\,{\scriptsize ($+8.0$)} & 43.2\,{\scriptsize ($0.0$)} & 69.4\,{\scriptsize ($+7.5$)} & 83.6\,{\scriptsize ($+0.9$)} & 74.1\,{\scriptsize ($+0.4$)} & 62.9\,{\scriptsize ($+0.3$)} & 93.9\,{\scriptsize ($-0.2$)}\\
+ Directed correction & 75.4\,{\scriptsize ($+6.1$)} & 57.9\,{\scriptsize ($0.0$)} & 64.4\,{\scriptsize ($+21.2$)} & 69.4\,{\scriptsize ($0.0$)} & 83.6\,{\scriptsize ($0.0$)} & 74.1\,{\scriptsize ($0.0$)} & 84.5\,{\scriptsize ($+21.7$)} & 93.9\,{\scriptsize ($0.0$)}\\
\bottomrule
\end{tabular*}
\par\vspace{4pt}
\begin{tabular*}{\textwidth}{@{\extracolsep{\fill}}l*{7}{l}@{}}
\toprule
\textbf{Potsdam} & mIoU & \shortstack[l]{Clutter\\(background)} & \shortstack[l]{Impervious\\surface} & Building & \shortstack[l]{Low\\vegetation} & Tree & Car\\
\midrule
Direct & 54.4 & 12.8 & 63.8 & 83.0 & 52.9 & 35.4 & 78.7\\
+ Foreground expansion & 56.7\,{\scriptsize ($+2.2$)} & 21.7\,{\scriptsize ($+8.9$)} & 72.4\,{\scriptsize ($+8.6$)} & 78.9\,{\scriptsize ($-4.1$)} & 55.0\,{\scriptsize ($+2.1$)} & 30.1\,{\scriptsize ($-5.4$)} & 82.0\,{\scriptsize ($+3.2$)}\\
+ Background expansion & 56.8\,{\scriptsize ($+0.1$)} & 22.5\,{\scriptsize ($+0.8$)} & 72.4\,{\scriptsize ($+0.1$)} & 78.9\,{\scriptsize ($0.0$)} & 55.0\,{\scriptsize ($0.0$)} & 30.1\,{\scriptsize ($0.0$)} & 82.0\,{\scriptsize ($0.0$)}\\
+ Foreground selection & 58.9\,{\scriptsize ($+2.1$)} & 22.4\,{\scriptsize ($-0.1$)} & 76.2\,{\scriptsize ($+3.7$)} & 84.5\,{\scriptsize ($+5.7$)} & 56.2\,{\scriptsize ($+1.1$)} & 32.2\,{\scriptsize ($+2.1$)} & 82.0\,{\scriptsize ($0.0$)}\\
+ Directed correction & 63.2\,{\scriptsize ($+4.2$)} & 22.4\,{\scriptsize ($0.0$)} & 77.4\,{\scriptsize ($+1.2$)} & 86.5\,{\scriptsize ($+1.9$)} & 60.7\,{\scriptsize ($+4.5$)} & 50.0\,{\scriptsize ($+17.8$)} & 82.0\,{\scriptsize ($0.0$)}\\
\bottomrule
\end{tabular*}
}
\par\vspace{6pt}
\end{table*}

\subsection{Class-wise Results}
Table~\ref{tab:complete-classes-a} reports the IoU for each output class. Buildings have the highest foreground IoU on Potsdam, Vaihingen, UDD5, and UAVid, with scores ranging from 86.5\% to 92.0\%. Water IoU is 58.0\% on LoveDA and 71.7\% on OpenEarthMap, reaching 93.9\% on VDD.

On iSAID, planes and storage tanks reach 73.8\% and 71.4\% IoU, respectively. Baseball diamonds and basketball courts reach 0.0\% and 3.1\% IoU. On OpenEarthMap, bareland IoU is 0.3\%.

\enlargethispage{\baselineskip}
\vspace{-3pt}
\subsection{Component Ablations on VDD and Potsdam}
Table~\ref{tab:additional-component-ablations} starts with the initial queries and semantic responses weighted by concept presence. We add the foreground candidate phrases, then use the LLM-generated background phrase set. Foreground phrase selection uses the selected phrase sets in Table~\ref{tab:inventory-all}. Directed correction then updates the initial labels using those same selected phrases. All stages share the background floor and class score computation.

Foreground expansion raises mIoU on Potsdam from 54.4\% to 56.7\%. At this step, road IoU on VDD falls from 64.6\% to 57.8\%, and tree IoU on Potsdam falls from 35.4\% to 30.1\%. Adding the background phrases raises background IoU on VDD from 42.3\% to 49.9\%. Foreground phrase selection then adds 2.4 points on VDD and 2.1 on Potsdam, averaging 2.3 points. After screening, water IoU on VDD drops from 94.1\% to 93.9\%, and clutter IoU on Potsdam drops from 22.5\% to 22.4\%. From the initial queries to the selected phrase sets, mIoU rises from 63.9\% to 69.3\% on VDD and from 54.4\% to 58.9\% on Potsdam, an average gain of 4.9 percentage points.

Directed correction adds 6.1 mIoU points on VDD and 4.2 on Potsdam. Roof IoU rises from 62.9\% to 84.5\%, and tree IoU from 32.2\% to 50.0\%. Correction mislabels an additional 3.3 million facade pixels as roof and 3.8 million low vegetation pixels as tree. Despite these errors, facade IoU rises from 43.2\% to 64.4\%, and low vegetation IoU rises from 56.2\% to 60.7\%. Figure~\ref{fig:conflict-correction} illustrates both corrections.

\begin{table*}[!t]
\centering
\small
\setlength{\tabcolsep}{3pt}
\setlength{\aboverulesep}{0.25ex}
\setlength{\belowrulesep}{0.25ex}
\renewcommand{\arraystretch}{1.04}
\caption{Sensitivity to phrase selection and minimum sample counts on six datasets under Eq.~\eqref{eq:correction}, in mIoU (\%). Bold marks the shared settings used in the main comparison.}
\label{tab:selection-sensitivity}
\begin{tabular*}{\textwidth}{@{\extracolsep{\fill}}lrrrrrrr@{}}
\toprule
Parameter & Value & Potsdam & Vaihingen & OpenEarthMap & UDD5 & VDD & UAVid\\
\midrule
$\tau_{\mathrm{own}}$ & 0.05 & 61.3 & 67.1 & 46.5 & 74.5 & 74.9 & 69.2\\
 & 0.1 & 61.3 & 67.1 & 46.5 & 74.5 & 74.9 & 69.2\\
 & 0.15 & 61.4 & 66.7 & 46.5 & 74.5 & 74.9 & 69.2\\
 & \textbf{0.2} & \textbf{63.2} & \textbf{67.7} & \textbf{46.4} & \textbf{74.5} & \textbf{75.4} & \textbf{69.0}\\
 & 0.25 & 63.2 & 67.7 & 45.6 & 74.5 & 75.6 & 69.0\\
 & 0.3 & 63.2 & 67.7 & 45.6 & 75.0 & 75.6 & 68.9\\
 & 0.4 & 62.7 & 67.7 & 45.5 & 75.0 & 75.6 & 67.4\\
\midrule
$\rho_{\mathrm{sep}}$ & 1 & 63.7 & 67.7 & 46.7 & 74.5 & 75.4 & 69.0\\
 & 1.25 & 63.7 & 67.7 & 47.4 & 74.5 & 75.4 & 69.0\\
 & \textbf{1.5} & \textbf{63.2} & \textbf{67.7} & \textbf{46.4} & \textbf{74.5} & \textbf{75.4} & \textbf{69.0}\\
 & 1.75 & 63.2 & 67.7 & 47.7 & 74.5 & 75.4 & 69.0\\
 & 2 & 63.2 & 65.6 & 48.2 & 74.5 & 75.4 & 69.0\\
 & 2.5 & 63.2 & 65.6 & 48.2 & 74.5 & 72.5 & 69.0\\
 & 3 & 63.2 & 65.6 & 48.1 & 74.5 & 72.5 & 69.0\\
\midrule
$n_{\mathrm{core}}$ & 1 or 8 or 16 & 63.2 & 67.7 & 46.1 & 74.5 & 75.4 & 69.0\\
 & \textbf{32} or 64 or 128 & \textbf{63.2} & \textbf{67.7} & \textbf{46.4} & \textbf{74.5} & \textbf{75.4} & \textbf{69.0}\\
\midrule
$n_{\mathrm{cal}}$ & \textbf{1} & \textbf{63.2} & \textbf{67.7} & \textbf{46.4} & \textbf{74.5} & \textbf{75.4} & \textbf{69.0}\\
 & 8 & 63.1 & 67.7 & 46.6 & 74.5 & 75.4 & 69.0\\
 & 32 & 63.1 & 67.7 & 46.4 & 74.5 & 75.4 & 69.0\\
 & 64 & 62.9 & 67.7 & 46.4 & 74.5 & 75.4 & 69.0\\
 & 128 & 62.8 & 67.8 & 47.0 & 74.5 & 75.4 & 69.0\\
 & 256 & 63.0 & 67.7 & 47.7 & 74.5 & 75.2 & 69.0\\
 & 512 & 63.0 & 67.6 & 47.4 & 74.5 & 75.4 & 69.0\\
\bottomrule
\end{tabular*}

\par\vspace{12pt}
{
\centering
\includegraphics[width=.75\textwidth,height=.48\textheight,keepaspectratio]{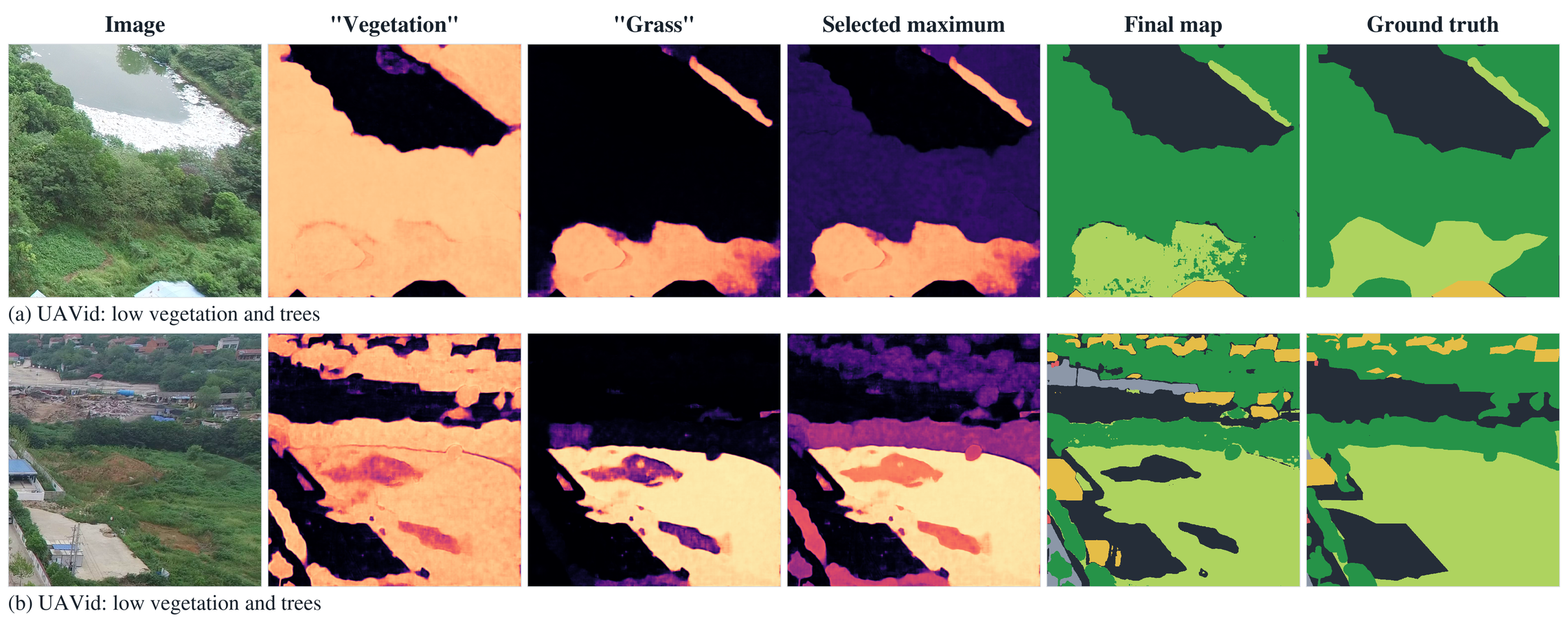}
\methodfigurecaption{On UAVid, vegetation responds to both trees and low vegetation, while grass concentrates on low vegetation. The low-vegetation class score (Selected maximum) is the maximum response over the selected phrases for that class. Heatmaps range from 0 (dark) to 1 (light).}
\label{fig:phrase-screening}
}
\end{table*}

\subsection{Parameter Sensitivity}
\label{sec:sensitivity}
Table~\ref{tab:selection-sensitivity} varies one setting at a time on six complete evaluation sets. Each dataset uses the same background floor as in the main comparison. These runs apply the correction threshold before introducing the requirement that the proposed target rank second. We recompute the selected phrase sets and correction rules affected by each setting. For each dataset, the three settings $n_{\mathrm{core}}\in\{1,8,16\}$ yield identical mIoU values, as do the three settings $n_{\mathrm{core}}\in\{32,64,128\}$. On OpenEarthMap, raising $n_{\mathrm{core}}$ from 16 to 32 changes mIoU from 45.9 to 46.2\%. On VDD, increasing $\rho_{\mathrm{sep}}$ from 1.5 to 2.5 reduces mIoU from 75.5 to 72.8\%.

\twocolumn[{\begin{minipage}{\textwidth}
\centering
{
\centering
\includegraphics[width=.75\textwidth,height=.72\textheight,keepaspectratio]{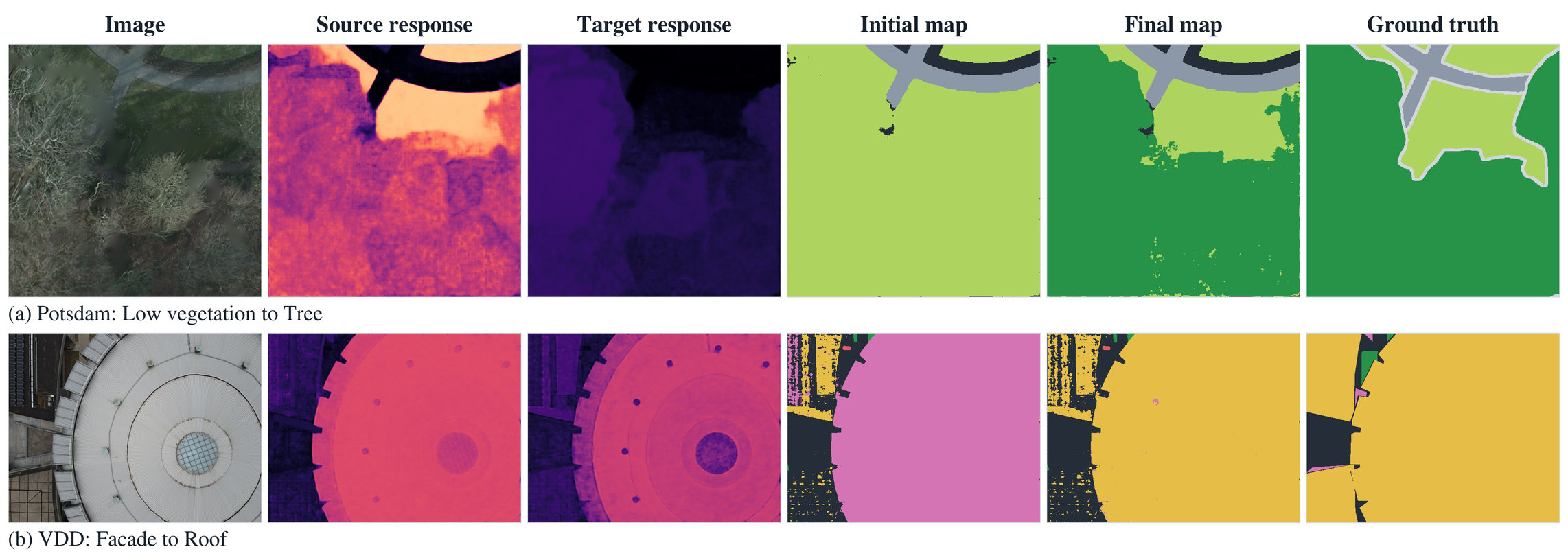}
\methodfigurecaption{Directed correction on Potsdam and VDD. The source and target response maps show the class scores $F_a$ and $F_b$. The Initial map and Final map panels show the effect of correction with the same selected phrase set. Heatmaps range from 0 (dark) to 1 (light).}
\label{fig:conflict-correction}
}
\par\vspace{3pt}
{
\includegraphics[width=.71\textwidth]{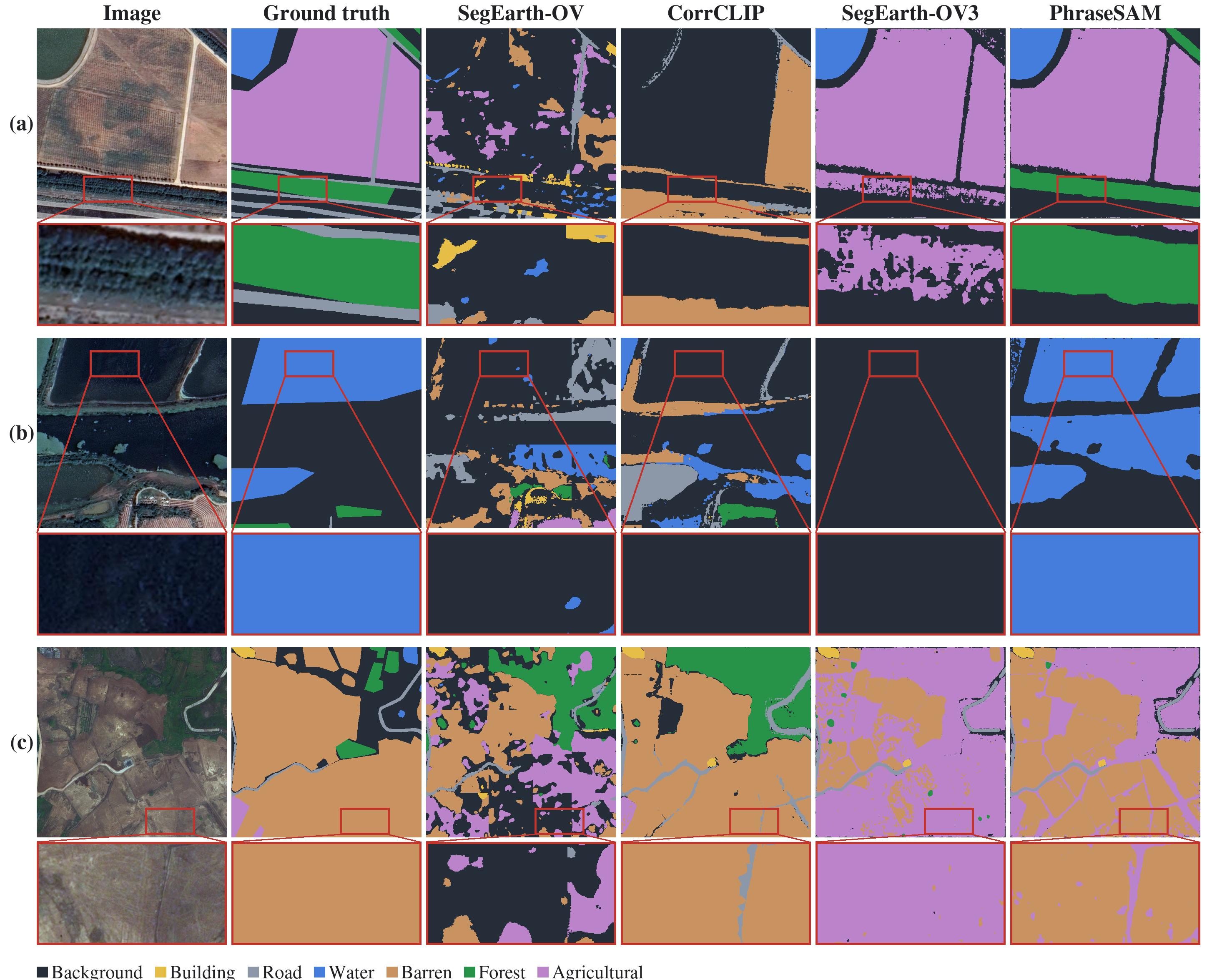}
\methodfigurecaption{Qualitative comparison on LoveDA. In (a), PhraseSAM recovers forest pixels that SegEarth-OV3 labels as farmland. In (b), PhraseSAM recovers water pixels labeled as background by SegEarth-OV3. In (c), PhraseSAM recovers barren land that SegEarth-OV3 labels as farmland. Pixel agreement inside the boxes rises from 18.5\% to 78.1\% in (a), 0.0\% to 100.0\% in (b), 0.7\% to 92.2\% in (c).}
\label{fig:qualitative-loveda}
}
\par\vspace{5pt}\end{minipage}}]

\begingroup
\setlength{\baselineskip}{0.955\baselineskip}
\subsection{Phrase Selection and Background Coverage}
Table~\ref{tab:inventory-all} lists all foreground and background candidate phrases alongside the selected phrases for the eight datasets under their assigned output classes. On UAVid, \emph{grass}, \emph{lawn}, \emph{short grass}, \emph{low plants}, and \emph{ground cover} represent low vegetation. Figure~\ref{fig:phrase-screening} compares these responses with the initial \emph{vegetation} query. The LLM-generated background phrase set remains fixed throughout foreground screening.

\subsection{Directed Corrections and Class Assignments}
On VDD, the number of ground-truth roof pixels assigned to facade decreases from 38.0 million in the initial label map to 6.9 million after correction. On Potsdam, the corresponding change for tree pixels assigned to low vegetation is from 41.7 million to 24.4 million.

Figure~\ref{fig:conflict-correction} shows low-vegetation-to-tree correction on Potsdam and facade-to-roof correction on VDD.

\subsection{Qualitative Comparisons on Eight Datasets}
Figures~\ref{fig:qualitative-loveda} through~\ref{fig:qualitative-uavid} compare PhraseSAM with SegEarth-OV, CorrCLIP, and SegEarth-OV3. For each dataset, we show three examples. Each example shows the complete evaluated input and an enlarged view of the same boxed region across all columns. We select examples in which PhraseSAM agrees more closely with the ground truth than SegEarth-OV3 in the box and across the complete image. All masks use the same dataset palette. In the Potsdam and Vaihingen figures, ignored boundary pixels take the color of the nearest labeled pixel for display.

\par\endgroup

\begin{table*}[!tp]
\centering
\fontsize{6.8}{7.3}\selectfont
\setlength{\tabcolsep}{2pt}
\renewcommand{\arraystretch}{0.96}
\setlength{\aboverulesep}{0.25ex}
\setlength{\belowrulesep}{0.25ex}
\caption{Candidate and selected phrases for the eight OVR-S8 datasets. Background phrases are assigned to the residual label.}
\label{tab:inventory-all}
\label{tab:inventory-loveda-potsdam}
\label{tab:selected-inventory-a}
\label{tab:inventory-vaihingen-uavid}
\label{tab:inventory-oem-udd5}
\label{tab:selected-inventory-b}
\label{tab:inventory-cont}
\label{tab:inventory-isaid-vdd}
\begin{tabular}{>{\raggedright\arraybackslash}p{.125\textwidth}>{\raggedright\arraybackslash}p{.425\textwidth}>{\raggedright\arraybackslash}p{.425\textwidth}}
\toprule
Output class & Candidate phrases & Selected phrases\\
\midrule
\multicolumn{3}{l}{\textit{LoveDA}}\\
Background & background, miscellaneous land, snow, sky, fog, construction debris & background, miscellaneous land, snow, sky, fog, construction debris\\
Building & building, house, roof, rooftop, building rooftop & building, house, building rooftop\\
Road & road, street, asphalt road, paved road, roadway & road, paved road, street, asphalt road\\
Water & water, river, lake, pond, water body & water, water body\\
Barren & barren, bare soil, bare ground, exposed soil, soil, barren land, dirt ground & barren, bare soil, dirt ground, soil\\
Forest & forest, tree, woodland, dense forest, tree canopy, dense trees & forest, tree, dense forest, dense trees\\
Agricultural & agricultural, farmland, crop field, cultivated field, agricultural land, planted field, cropland & agricultural, cropland, farmland, cultivated field\\
\midrule
\multicolumn{3}{l}{\textit{Potsdam}}\\
Impervious surface & road, path, sidewalk, paved surface, pavement & road, pavement\\
Building & building, house, roof, rooftop, building rooftop & building, house, roof, rooftop, building rooftop\\
Low vegetation & grass, low vegetation, lawn, ground grass, short grass, low plants, ground cover & grass, lawn, ground grass, short grass, ground cover\\
Tree & tree, tree canopy, individual tree, woody vegetation, forest, woodland, tall tree & tree, individual tree, tall tree\\
Car & car, vehicle, automobile, parked car, moving car & car, vehicle, automobile, parked car, moving car\\
Clutter & clutter, scattered, sand, snow, sky, sports court, river bank, swimming pool, shipping container & clutter, scattered, sand, snow, sky, sports court, river bank, swimming pool, shipping container\\

\midrule
\multicolumn{3}{l}{\textit{Vaihingen}}\\
Impervious surface & road, path, sidewalk, paved surface, pavement & road, paved surface, pavement\\
Building & building, house, roof, rooftop, building rooftop & building, house, roof, rooftop, building rooftop\\
Low vegetation & grass, low vegetation, lawn, ground grass, short grass, low plants, ground cover & grass, lawn, ground grass, short grass, ground cover\\
Tree & tree, tree canopy, individual tree, woody vegetation, forest, woodland, tall tree & tree, individual tree, tall tree\\
Car & car, vehicle, automobile, parked car, moving car & car, vehicle, automobile, parked car, moving car\\
Clutter & clutter, sand, snow, sky, sports court, river bank, swimming pool, shipping container & clutter, sand, snow, sky, sports court, river bank, swimming pool, shipping container\\
\midrule
\multicolumn{3}{l}{\textit{UAVid}}\\
Background & background, other, cluttered ground, sand, sky, parking lot, plaza, sidewalk, sports court, bare soil, river, lake, driveway, loading bay, bus stop, utility pole, streetlight, lamp post, fence, mailbox, cloud, fog, smoke & background, other, cluttered ground, sand, sky, parking lot, plaza, sidewalk, sports court, bare soil, river, lake, driveway, loading bay, bus stop, utility pole, streetlight, lamp post, fence, mailbox, cloud, fog, smoke\\
Building & building, house, roof, rooftop, building rooftop & building, house, roof, rooftop, building rooftop\\
Road & road, street, asphalt road, paved road, roadway & road, street, asphalt road, paved road, roadway\\
Car & car, vehicle, automobile, parked car, moving car & car, vehicle, automobile, parked car, moving car\\
Tree & tree, tree canopy, individual tree, woody vegetation, forest, woodland, tall tree & tree, individual tree, tall tree\\
Low vegetation & vegetation, grass, low vegetation, lawn, short grass, low plants, ground cover & grass, lawn, short grass, low plants, ground cover\\
Human & human, person, pedestrian, people & human, person, pedestrian, people\\
\midrule
\multicolumn{3}{l}{\textit{OpenEarthMap}}\\
Background & background, snow, bridge, cloud, smoke, cloud shadow, mist band & background, snow, bridge, cloud, smoke, cloud shadow, mist band\\
Bareland & bareland, bare soil, bare ground, exposed soil, soil, barren land & bareland\\
Grass & grass, grassland, green field, grassy ground, lawn & grass, green field, grassy ground, lawn\\
Pavement & pavement, paved area, concrete ground, paved ground, concrete surface & pavement, paved area, concrete ground, paved ground, concrete surface\\
Road & road, street, asphalt road, paved road, roadway & road, asphalt road, paved road, roadway\\
Tree & tree, tree canopy, individual tree, woody vegetation, forest, woodland, tall tree & tree, tree canopy, individual tree, forest\\
Water & water, river, lake, pond, water body & water, river, lake, water body\\
Cropland & cropland, farmland, crop field, cultivated field, agricultural land, planted field & cropland, farmland, crop field, cultivated field, agricultural land\\
Building & building, house, roof, rooftop, building rooftop & building, house, roof, rooftop, building rooftop\\
\midrule
\multicolumn{3}{l}{\textit{UDD5}}\\
Vegetation & vegetation, tree, grass, shrubs, aquatic vegetation, lotus leaves & vegetation, tree, grass, shrubs\\
Building & building, house, roof, rooftop, building rooftop & building, house, roof, rooftop, building rooftop\\
Road & road, street, asphalt road, paved road, roadway & road, street, asphalt road, paved road, roadway\\
Vehicle & vehicle, car, automobile, truck, bus & vehicle, car, automobile\\
Background & background, sky, sports field, sand, sidewalk, parking lot, plaza, sports court, river bank, open water, parking bay, road shoulder, fence & background, sky, sports field, sand, sidewalk, parking lot, plaza, sports court, river bank, open water, parking bay, road shoulder, fence\\

\midrule
\multicolumn{3}{l}{\textit{iSAID}}\\
Background & background, Land Cover, snow, sand, railway track, roof, cloud, parking lot, plaza, sidewalk, natural ground, driveway, unclaimed pavement, manhole cover, street light, utility pole, guardrail, haze & background, Land Cover, snow, sand, railway track, roof, cloud, parking lot, plaza, sidewalk, natural ground, driveway, unclaimed pavement, manhole cover, street light, utility pole, guardrail, haze\\
Ship & ship, vessel, boat, cargo ship & ship, vessel, boat, cargo ship\\
Storage tank & storage tank, oil tank, cylindrical storage tank, industrial storage tank & storage tank, oil tank, cylindrical storage tank, industrial storage tank\\
Baseball diamond & baseball diamond, baseball field, baseball playing field & baseball diamond\\
Tennis court & tennis court, outdoor tennis court, tennis playing court & tennis court, outdoor tennis court, tennis playing court\\
Basketball court & basketball court, outdoor basketball court, basketball playing court & basketball court, basketball playing court\\
Running track & running track, athletic track, athletics track, track and field ground & running track, athletic track\\
Bridge & bridge, road bridge, river bridge, overpass & bridge, road bridge, river bridge, overpass\\
Large vehicle & large vehicle, heavy vehicle, large truck, large bus & large vehicle, large truck, large bus\\
Small vehicle & small vehicle, small car, passenger car, small road vehicle & small vehicle, small car, passenger car\\
Helicopter & helicopter, rotorcraft & helicopter\\
Swimming pool & swimming pool, outdoor swimming pool, rectangular swimming pool & swimming pool, outdoor swimming pool, rectangular swimming pool\\
Roundabout & roundabout, traffic circle, circular road junction & roundabout\\
Soccer field & soccer field, football field, soccer pitch, football pitch & soccer field, football field, soccer pitch, football pitch\\
Plane & plane, airplane, aeroplane, fixed wing aircraft & plane, airplane, aeroplane, fixed wing aircraft\\
Harbor & harbor, harbour, seaport, dockyard & harbor, harbour\\
\midrule
\multicolumn{3}{l}{\textit{VDD}}\\
Background & background, other, sand, footpath, sky, parking lot, plaza, sidewalk, sports court, dirt path & background, other, sand, footpath, sky, parking lot, plaza, sidewalk, sports court, dirt path\\
Facade & facade, wall, exterior wall, building facade, vertical building wall, exterior walls & facade, wall, exterior wall, building facade, vertical building wall, exterior walls\\
Road & road, street, asphalt road, paved road, roadway & road\\
Vegetation & vegetation, tree, grass, shrubs, plant cover & vegetation, tree, shrubs\\
Vehicle & vehicle, car, automobile, truck, bus & vehicle, car, automobile\\
Roof & roof, rooftop, building rooftop, roof surface, flat roof, sloped roof & roof, rooftop, building rooftop, roof surface, flat roof\\
Water & water, river, lake, pond, water body & water, river, lake, pond, water body\\

\bottomrule
\end{tabular}
\end{table*}

\begin{figure*}[!tp]
\centering
{
\includegraphics[width=.71\textwidth]{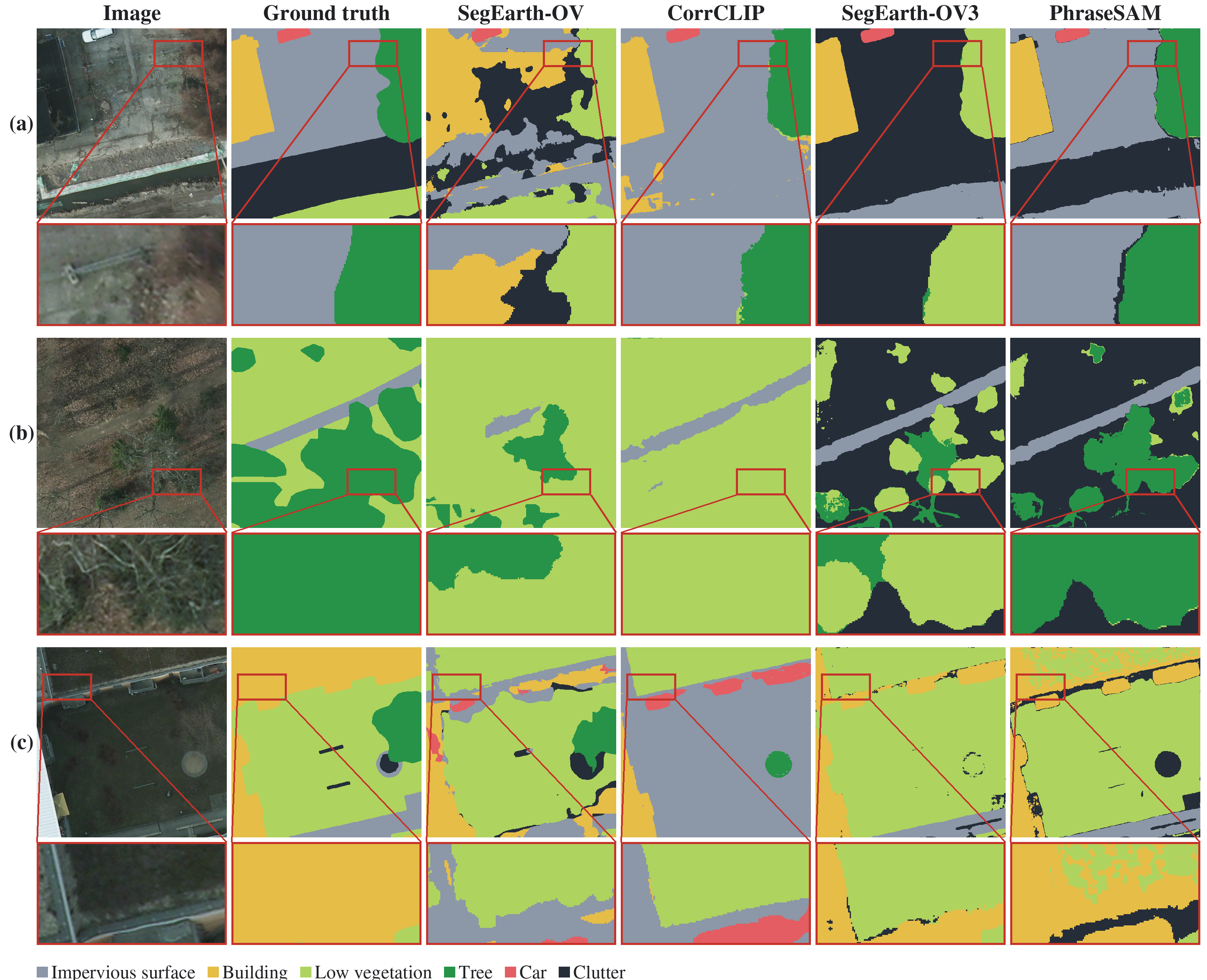}
\methodfigurecaption{Qualitative comparison on Potsdam. In (a), PhraseSAM recovers impervious surface pixels labeled as clutter by SegEarth-OV3. In (b), PhraseSAM recovers the lower-right tree canopy, which SegEarth-OV3 partly labels as low vegetation. In (c), PhraseSAM recovers building pixels labeled as low vegetation by SegEarth-OV3. Pixel agreement inside the boxes rises from 0.5\% to 98.2\% in (a), 13.9\% to 82.5\% in (b), 32.7\% to 69.4\% in (c).}
\label{fig:qualitative-potsdam}
}
\par\vspace{7pt}
{
\includegraphics[width=.71\textwidth]{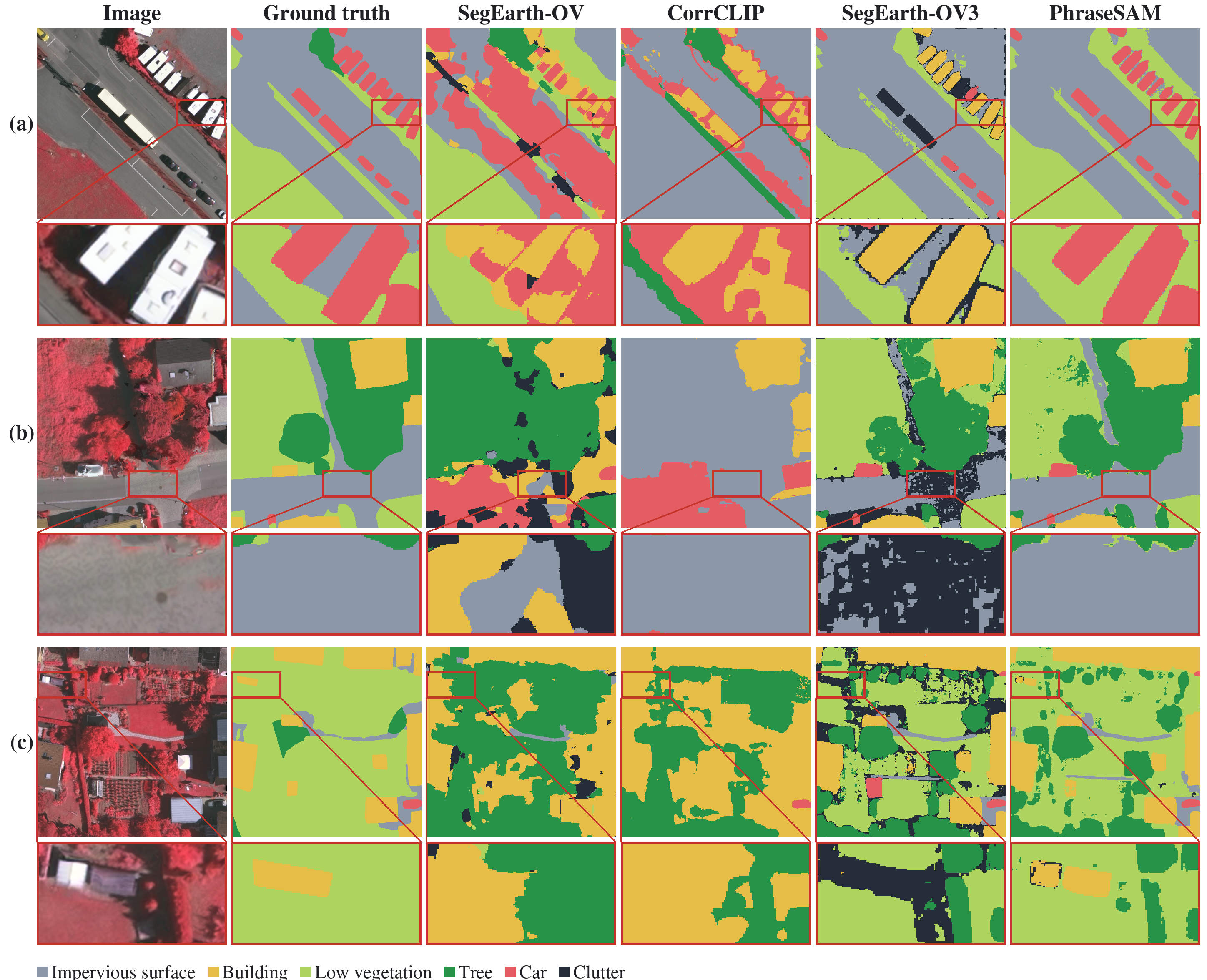}
\methodfigurecaption{Qualitative comparison on Vaihingen. In (a), PhraseSAM recovers parked vehicles that SegEarth-OV3 labels as building. In (b), PhraseSAM recovers impervious surface pixels labeled as clutter by SegEarth-OV3. In (c), PhraseSAM recovers building pixels labeled as clutter by SegEarth-OV3. Pixel agreement inside the boxes rises from 48.8\% to 88.3\% in (a), 22.5\% to 98.1\% in (b), 54.3\% to 80.7\% in (c).}
\label{fig:qualitative-vaihingen}
}
\end{figure*}

\begin{figure*}[!tp]
\centering
{
\includegraphics[width=.71\textwidth]{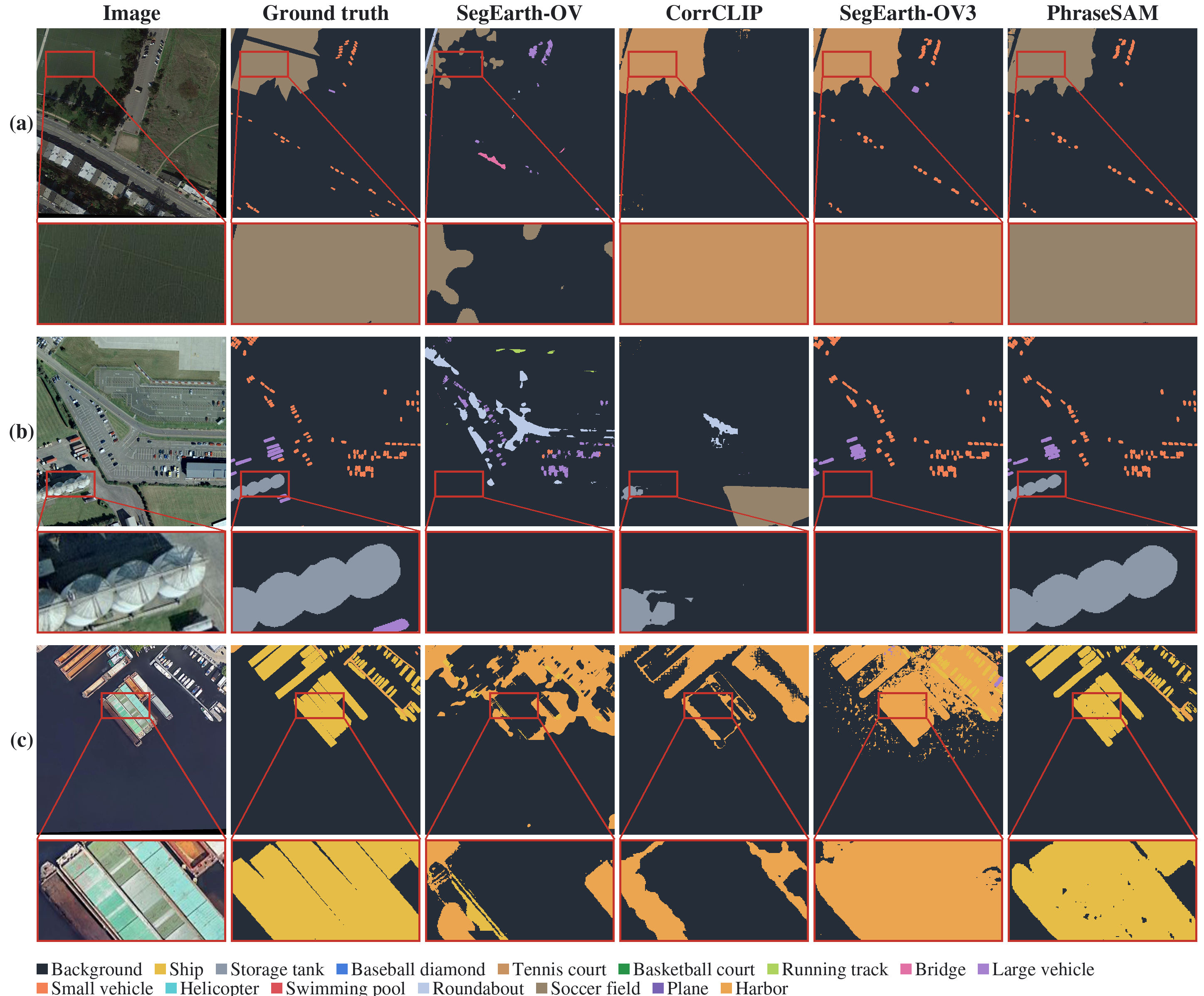}
\methodfigurecaption{Qualitative comparison on iSAID. In (a), PhraseSAM recovers soccer field pixels labeled as tennis court by SegEarth-OV3. In (b), PhraseSAM recovers storage tank pixels labeled as background by SegEarth-OV3. In (c), PhraseSAM recovers ship pixels labeled as harbor by SegEarth-OV3. Pixel agreement inside the boxes rises from 0.1\% to 99.2\% in (a), 58.7\% to 96.0\% in (b), 7.8\% to 94.1\% in (c).}
\label{fig:qualitative-isaid}
}
\par\vspace{7pt}
{
\includegraphics[width=.71\textwidth]{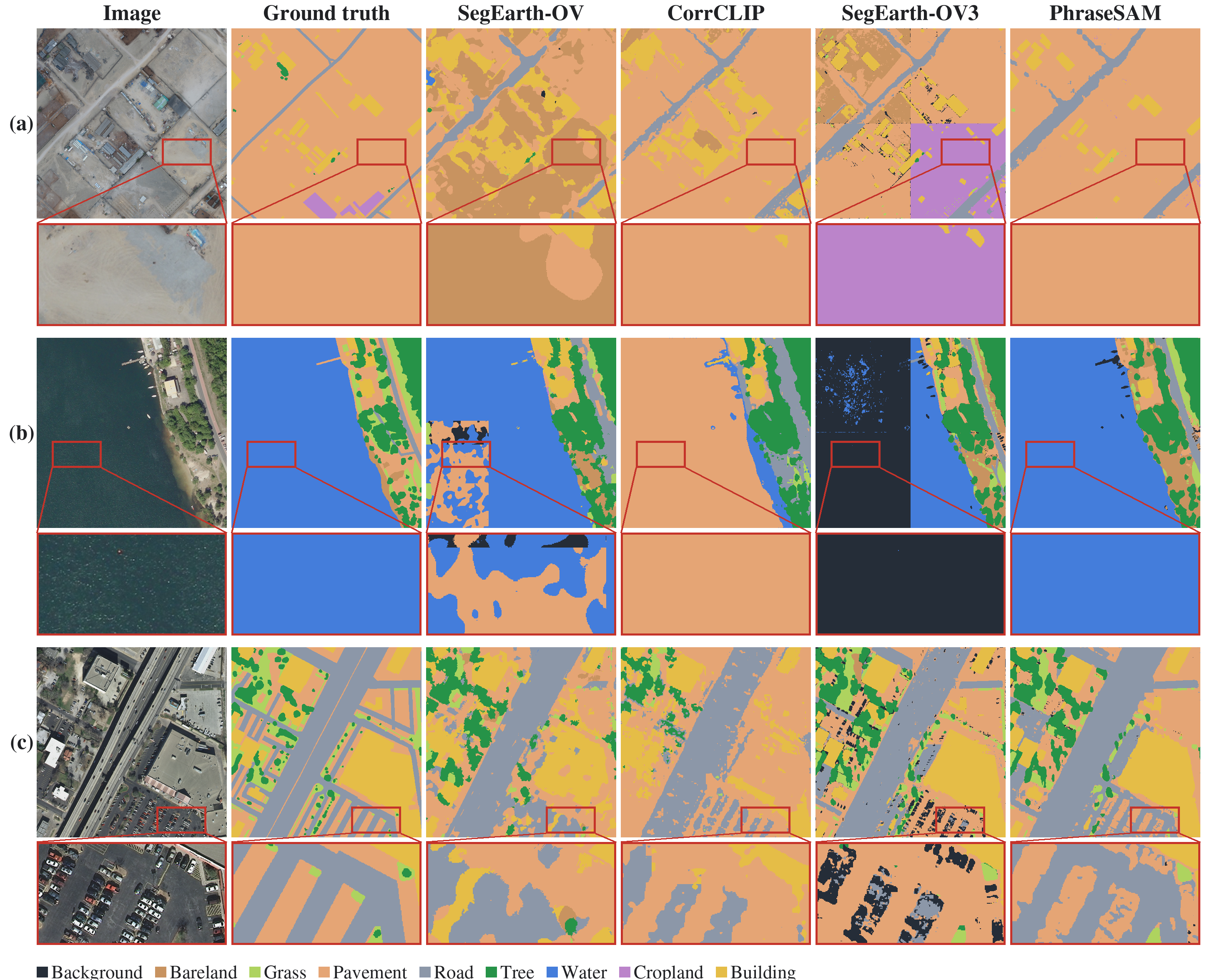}
\methodfigurecaption{Qualitative comparison on OpenEarthMap. In (a), PhraseSAM recovers pavement pixels labeled as cropland by SegEarth-OV3. In (b), PhraseSAM recovers water pixels labeled as background by SegEarth-OV3. In (c), PhraseSAM recovers road pixels labeled as pavement by SegEarth-OV3. Pixel agreement inside the boxes rises from 0.0\% to 100.0\% in (a), 0.0\% to 100.0\% in (b), 24.0\% to 87.1\% in (c).}
\label{fig:qualitative-openearthmap}
}
\end{figure*}

\begin{figure*}[!tp]
\centering
{
\includegraphics[width=.71\textwidth]{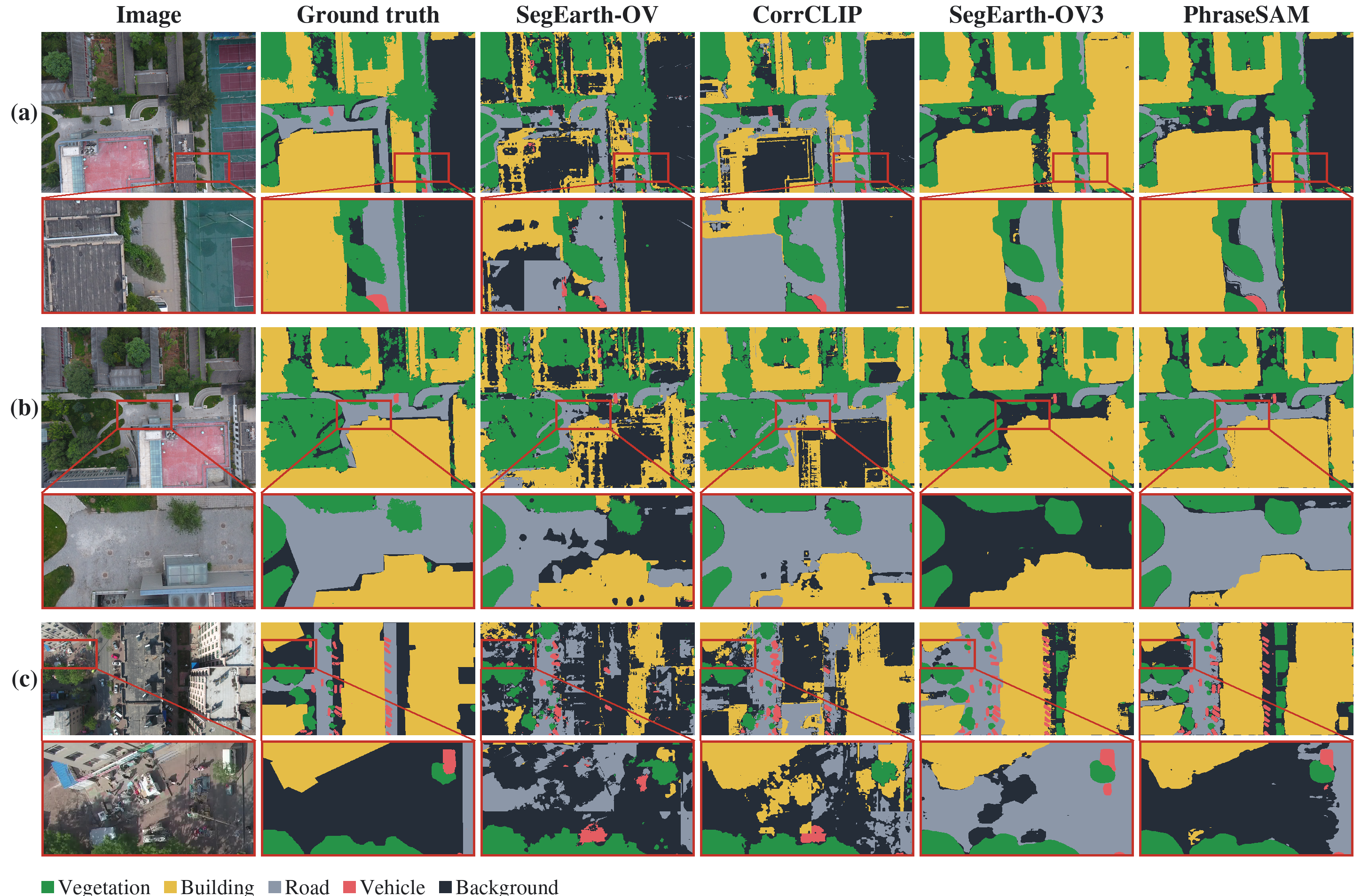}
\methodfigurecaption{Qualitative comparison on UDD5. In (a), PhraseSAM recovers background pixels labeled as building by SegEarth-OV3. In (b), PhraseSAM recovers road pixels labeled as background by SegEarth-OV3. In (c), PhraseSAM labels cluttered ground as background, where SegEarth-OV3 predicts road. Pixel agreement inside the boxes rises from 63.8\% to 95.9\% in (a), 44.7\% to 86.9\% in (b), 41.0\% to 92.2\% in (c).}
\label{fig:qualitative-udd5}
}
\par\vspace{7pt}
{
\includegraphics[width=.71\textwidth]{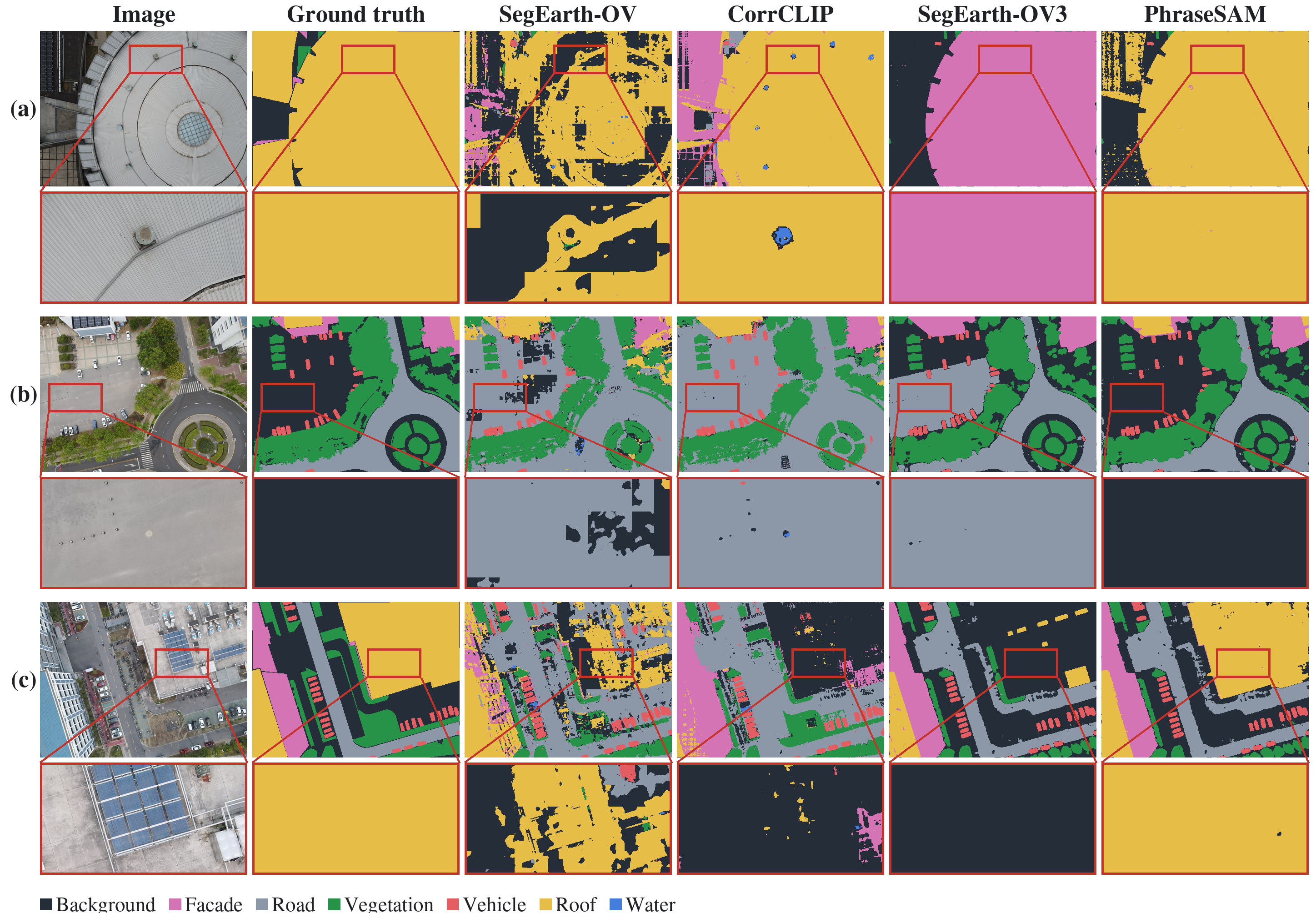}
\methodfigurecaption{Qualitative comparison on VDD. In (a), PhraseSAM recovers roof pixels labeled as facade by SegEarth-OV3. In (b), PhraseSAM recovers background pixels labeled as road by SegEarth-OV3. In (c), PhraseSAM recovers roof pixels labeled as background by SegEarth-OV3. Pixel agreement inside the boxes rises from 0.0\% to 100.0\% in (a), 0.0\% to 100.0\% in (b), 0.0\% to 99.9\% in (c).}
\label{fig:qualitative-vdd}
}
\end{figure*}

\twocolumn[{\begin{minipage}{\textwidth}
\centering
{
\includegraphics[width=.71\textwidth]{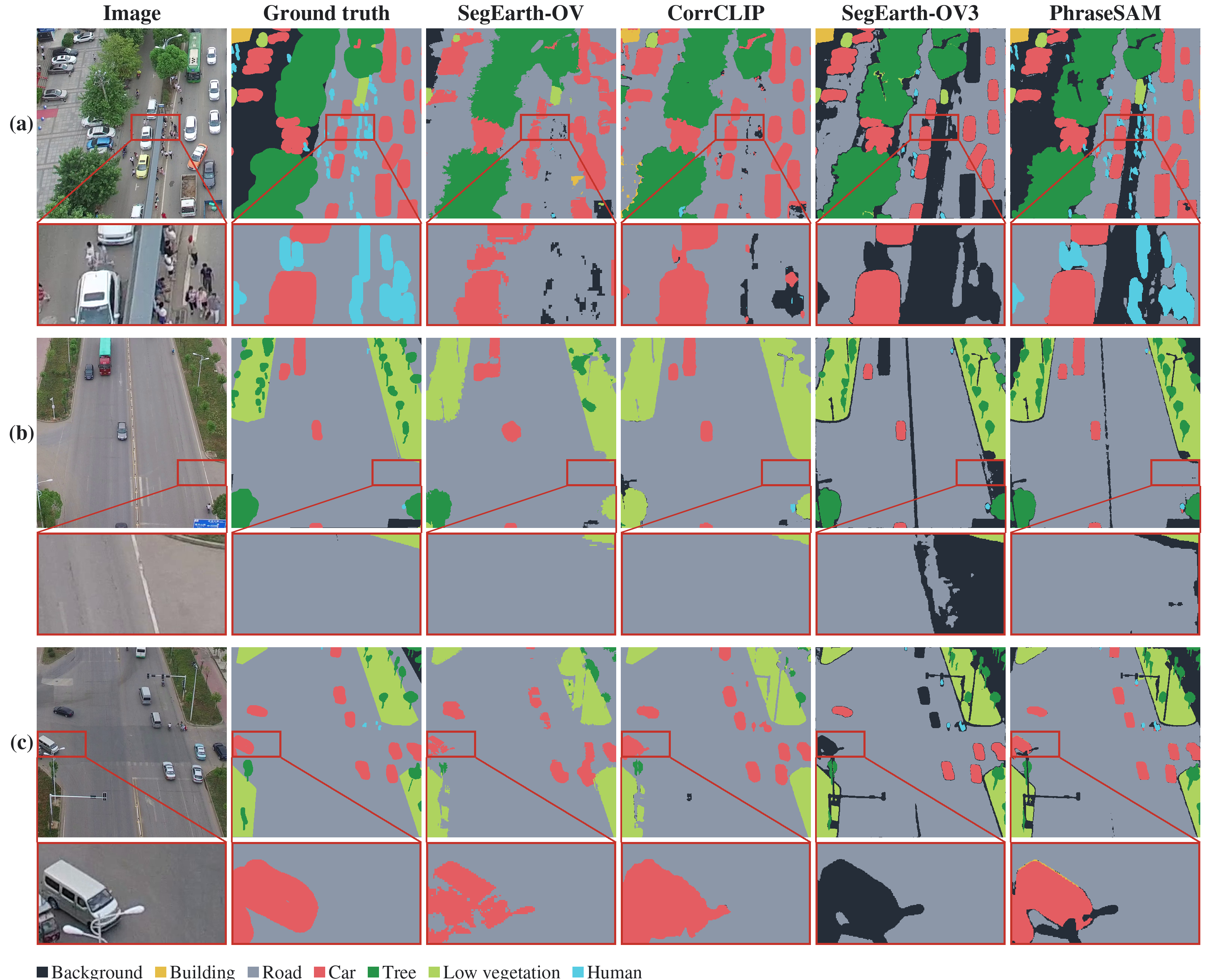}
\methodfigurecaption{Qualitative comparison on UAVid. In (a), PhraseSAM recovers pedestrians that SegEarth-OV3 labels as background. In (b), PhraseSAM recovers road pixels labeled as background by SegEarth-OV3. In (c), PhraseSAM identifies a vehicle that SegEarth-OV3 labels as background. Pixel agreement inside the boxes rises from 54.5\% to 70.9\% in (a), 63.9\% to 95.5\% in (b), 70.1\% to 91.8\% in (c).}
\label{fig:qualitative-uavid}
}
\par\vspace{\dbltextfloatsep}\end{minipage}}]
\section{Conclusion}
We presented PhraseSAM, a training-free framework for open-vocabulary remote-sensing segmentation with frozen SAM 3. The phrase expansion and selection stage combines foreground phrases selected for class support, separation, and complementary coverage with a background phrase set defined by the full class list. The correction stage identifies competing classes from mutual responses and uses calibration pixels to set correction thresholds. The selected phrase set and correction rules are fixed for inference, with each pixel receiving at most one correction. Across eight remote-sensing datasets, PhraseSAM achieves 60.5\% mean mIoU compared with 54.2\% for GLoP-SAM3, while the cumulative ablations measure the change as each stage is added on VDD and Potsdam.

\end{document}